\documentclass{article}
\usepackage{arxiv}
\usepackage{textcomp}
\usepackage[utf8]{inputenc} 
\usepackage[T1]{fontenc}    
\usepackage{hyperref}       
\usepackage{url}            
\usepackage{booktabs}       
\usepackage{amsfonts}       
\usepackage{amsmath}
\usepackage{nicefrac}     
\usepackage{microtype}      
\usepackage{lipsum}		
\usepackage{graphicx}
\usepackage{diagbox}
\usepackage{makecell}
\usepackage{float}
\usepackage{cleveref}
\usepackage[symbol]{footmisc}
\usepackage{geometry}
\usepackage[square,numbers]{natbib}
\usepackage{doi}
\usepackage{titling}
\usepackage[section]{placeins}
\usepackage{xcolor}
\usepackage{needspace}
\creflabelformat{equation}{#2\textup{#1}#3}

\title{Establishing Deep InfoMax as an effective self-supervised learning methodology in materials informatics}

\author{\textbf{Michael Moran, Vladimir V. Gusev*, Michael W. Gaultois,} \\ \textbf{Dmytro Antypov, Matthew J. Rosseinsky*} \\
\\
Leverhulme Research Centre for Functional Materials Design \\
University of Liverpool \\
Liverpool, United Kingdom \\ \\
Department of Chemistry \\
University of Liverpool \\
Liverpool, United Kingdom \\
M.J.Rosseinsky@liverpool.ac.uk\\ \\
Department of Computer Science \\
Department of Chemistry, University of Liverpool \\
Liverpool, United Kingdom \\
Vladimir.Gusev@liverpool.ac.uk}

\renewcommand{\shorttitle}{Deep InfoMax for materials informatics}

\hypersetup{
pdftitle={Establishing Deep InfoMax as an effective self-supervised Learning methodology in Materials Informatics},
pdfauthor={Michael~Moran},
pdfkeywords={point set, Transformer, Self-attention, Crystallography, Machine learning, Deep learning, Property prediction, materials informatics, Self-supervised Learning, Site-Net, Deep-InfoMax},
}

\begin{document}
\maketitle


\begin{abstract}

The scarcity of property labels remains a key challenge in materials informatics, whereas materials data without property labels are abundant in comparison. By pretraining supervised property prediction models on self-supervised tasks that depend only on the "intrinsic information" available in any Crystallographic Information File (CIF), there is potential to leverage the large amount of crystal data without property labels to improve property prediction results on small datasets. We apply Deep InfoMax as a self-supervised machine learning framework for materials informatics that explicitly maximises the mutual information between a point set (or graph) representation of a crystal and a vector representation suitable for downstream learning. This allows the pretraining of supervised models on large materials datasets without the need for property labels and without requiring the model to reconstruct the crystal from a representation vector. We investigate the benefits of Deep InfoMax pretraining implemented on the Site-Net architecture to improve the performance of downstream property prediction models with small amounts ($<10^3$) of data, a situation relevant to experimentally measured materials property databases. Using a property label masking methodology, where we perform self-supervised learning on larger supervised datasets and then train supervised models on a small subset of the labels, we isolate Deep InfoMax pretraining from the effects of distributional shift. We demonstrate performance improvements in the contexts of representation learning and transfer learning on the tasks of band gap and formation energy prediction. Having established the effectiveness of Deep InfoMax pretraining in a controlled environment, our findings provide a foundation for extending the approach to address practical challenges in materials informatics.

\end{abstract}

\pagebreak

\section{Introduction}
\label{sec:intro}

Materials Informatics has seen the development of many advanced transformers\cite{Cui2023, Site-Net} and graph models\cite{cgcnn,megnet} to perform supervised machine learning on crystal structures, i.e., on materials where both composition and structure are known.
These models work by breaking down a structure into its constituent atomic sites and the relationships between them.
In both of these paradigms the sites exchange information with each other to establish their context within the crystal, before being pooled into a fixed sized vector representation of the crystal that can be transformed into a property prediction.
This vector encodes all the crystal as points in a shared vector space and can from there be trivially used for property prediction and other tasks.

Supervised machine learning relies on property labels.
Property labels are the non-trival to obtain information about an input that you wish to predict using a machine learning model.
This is as opposed to what in this work we will call the intrinsic information in the input which is either directly present or can be trivially computed.
In the context of materials informatics and machine learning on crystal structures in particular, property labels are the physical properties of materials that require measurement or are difficult to computationally model.
The intrinsic information is the stoichiometry of the crystal; the geometry of the sites in the crystal structure. What we call the intrinsic information is everything that is immediately and directly computable from a Crystallographic Information File (CIF)\cite{Hall:es0164}.

Although transformers and graph models possess high expressiveness and the ability to capture complex chemical information, their performance in supervised learning is ultimately limited by the availability of data with property labels.
Data scarcity remains a significant challenge in materials science\cite{chang2022,NANDY2022100778}, as property specific data sets tend to be small and measurements can be noisy.
This is compounded for transformers and graph models as they often contain orders of magnitude more parameters than models like decision trees and simple multilayer perceptrons and, as such, are expected to be vulnerable to overtraining when there is a small amount of data.

When only small amounts of labelled data are available for a given property, additional information can be obtained from external sources.
An example of this is supervised transfer learning\cite{Hutchinson2017transferlearning}, in which a donor model is trained on a related property where property labels are more abundant, and a target model then makes use of the donor model to improve performance on a task where there is less data.
Whether through using predictions for this related property as a feature in the target model or by direct reuse of model parameters from the donor model. This can often lead to gains in performance on the target task if the properties are sufficiently correlated.
Although supervised transfer learning is often effective, it can be difficult in practice to predict which properties are sufficiently correlated enough for benefits to be obtained.

Self-supervised transfer learning presents an alternative to the supervised approach described above.
Unlike supervised learning, self-supervised learning does not require property labels, instead focusing on tasks that only include the intrinsic information in the input i.e the unlabelled data supervises itself.
In the context of materials informatics, these methodologies take advantage of intrinsic information from large databases of crystal structures with no property labels provided.
This information is then used to improve the performance of supervised models trained on small datasets by guiding learning through more generic chemical rules and understanding.
In contrast to supervised transfer learning, which relies on some other property being well-correlated with the target property, self-supervised transfer learning leverages intrinsic information about crystals available in any Crystallographic Information File (CIF).

Self-supervised transfer learning is expected to more readily generalise than any particular property label used for supervised transfer learning, because self-supervised tasks can cast a broader net on what knowledge to extract.
For transformers and graph models in particular, there is a strong motivation for self-supervised transfer learning as the step of aggregating the crystal into a useful vector likely has components that are shared between many tasks.
One example is capturing the geometric information of the crystal which is expected to be a necessary component of most property prediction tasks.
In this setup, we rely on a large unlabelled crystal database to teach a model what a crystal is, and this information can then be used to improve performance on downstream supervised tasks.
Through this methodology, the risk for overtraining can be reduced, as less of the models final behaviour depends on the small labelled dataset and more on picking up patterns found in known crystals.
At its most ambitious, large-scale 'foundation models'\cite{Schneider2024} can be trained on very large datasets combining many self-supervised tasks, acting as a general purpose starting point for supervised models.

In this work, we adapt Deep InfoMax\cite{hjelm2019learning,sun2020infograph} as a self-supervised learning framework for crystal transformers and crystal graphs that maximises mutual information between the constituents\footnote[3]{To avoid confusion between chemical elements, and the elements of a set, we refer to the elements of a set as the constituents of a set in this work} of the initial representation of the crystal used by the model and the produced feature vector.
We investigate the benefits of this approach for improving the performance of downstream property prediction models where only small amounts of property labels are available.
This is investigated through using the learned crystal representation vectors as input features to simple supervised property prediction models, and through transfer learning where the parameters from models trained with Deep InfoMax are used as the initial parameters for supervised models.
We also visualise the latent space of the learned representation, that is, the representation vector of the crystal learned by the model.
This allows for a qualitative assessment of the effects of Deep InfoMax training.
Unlike autoencoders\cite{Kingma_2019}, Deep InfoMax does not require reconstructing the original crystal as part of the pipeline.
Reconstructing crystals from vector representations without external information injection (i.e, the number of atoms in the unit cell and the unit cell parameters) remains an unsolved problem, so the lack of reconstruction requirements is an important advantage of our approach.

This self-supervised pretraining approach, where a model is first trained on a large unlabelled dataset, and the gathered insights are used to improve performance on small supervised datasets is normally performed with distinct datasets.
In this regime the small supervised dataset is unrelated to the large dataset used for pre-training.
Using distinct datasets is necessary for real world applications of self-supervised pre-training but in establishing the suitability of a technique the use of two distinct datasets introduces an undesirable extra degree of freedom, distributional shift, where the large dataset may simply not be relevant to the target dataset.
To mitigate this and isolate Deep InfoMaxes capabilities from distributional shift, we use the band gap and formation energy datasets from the Materials Project for both pretraining and supervised training.
Deep InfoMax models are trained on the entirety of each dataset (approximately $10^5$ samples) without using the property labels, and supervised training takes place on multiple randomly sampled subsets of each dataset.
This allows us to benchmark the effectiveness of Deep InfoMax on each task as a function of the amount of available data for supervised training, with well defined statistical resolution.


\section{Methods}

\subsection{Crystal transformers and graphs - self-supervised learning on permutation invariant functions}

In this work, we build the Deep InfoMax architecture on top of Site-Net\cite{Site-Net}, which has been adapted for self-supervised learning from its previous application for supervised property prediction.
Site-Net is a transformer architecture for crystals that performs complete self-attention on roughly cubic supercell representations of the crystal, similar to those used in simulations\cite{MILITZER20168}.
The supercell is represented as the set (an unordered list) of the local environments within it, where each local environment is itself the set of pairwise interaction features between the central site and every other site in the cell, including itself.
The sets of vectors representing the pairwise interaction features are aggregated into vectors representing the local environments, and the set of local environments is aggregated into a vector that represents the entire crystal.
Site-Net operates on cubic super-cells of a consistent size chosen as a hyper parameter, which acts as the main inductive bias.

\begin{equation}
\centering
z = f({c_{1}, ... , c_n}) = f(c_{\pi(1)}, ... , c_{\pi(n)})
\label{eq:inv}
\end{equation}

The fixed vector representations of a set is constructed through a permutation invariant function.
This permutation invariant function (\cref{eq:inv}) groups a variably sized set of vectors in a common basis ($c_i$) into a single fixed sized vector ($z$). For a set of size n, a permutation operator ($\pi$) can be applied that transforms indices 1 through n to some arbitrary new permutation and the result of the function does not change. 
In the context of Site-Net, this is the mapping of sets of vectors featurising the pairwise interactions to vectors featurising local environments, and vectors featurising the local environments to a vector featurising the crystal.
The encoder in a neural network that maps a variable size input to a fixed vector can be described as a parametric permutation invariant aggregation function.
This description of an encoder as a permutation invariant aggregation function is general and applies to set models, graph convolution models, and transformers.
This paradigm allows for the handling of crystals of arbitrary size and complexity and respects that there is no intrinsic ordering to the sites in a crystal or the pairwise interaction features in a local environment.

Permutation invariant aggregation functions, by their nature, are prone to information loss.
This is especially true when there are multiple layers of permutation invariant function between the input to the model and the representation.
To mitigate this, the permutation invariant functions can be optimised to capture the essential characteristics of the input for a given task.
One approach to solving this problem is through manual design of the functions based on known constraints on the input, without utilising data.
A particularly simple and successful example is to convert the set of atoms in a chemical composition into a vector of elemental fractions.
Elemental fraction vectors provide a bijective representation of the chemical composition.
That is, for every element fraction vector, there is exactly one composition, and for every composition there is one elemental fraction vector.
However, defining a similar function for structural information poses a significant challenge.

The other approach is to learn the parameters for the aggregation function from the data; this is the approach taken with Deep InfoMax.
Deep InfoMax maximises the mutual information between the crystal set ($c_i$) and the crystal representation ($z$).
With enough crystals to train on, it is possible for a model to learn the intrinsic structure of the crystals and to learn an efficient encoding that captures information relevant to the properties of interest.
Once the model is trained, the learned vectors representing the crystal, or the parameters of the encoder that generated them, can be used for downstream machine learning tasks.
The simplest use case is improving performance for supervised learning with small datasets. Although we use the Site-Net architecture in this work to implement Deep InfoMax for crystals, the methodology is fairly generic. Deep InfoMax can also be applied to graph models such as CGCNN, ALIGNN, and MegNet\cite{cgcnn,Choudhary2021,megnet} that fit the framework of sequences of permutation invariant aggregation functions.

\subsection{Intuition for mutual information maximisation - self-supervised learning with classifiers}

\begin{figure*}[!ht]
	\centering
	\includegraphics[width=\textwidth]{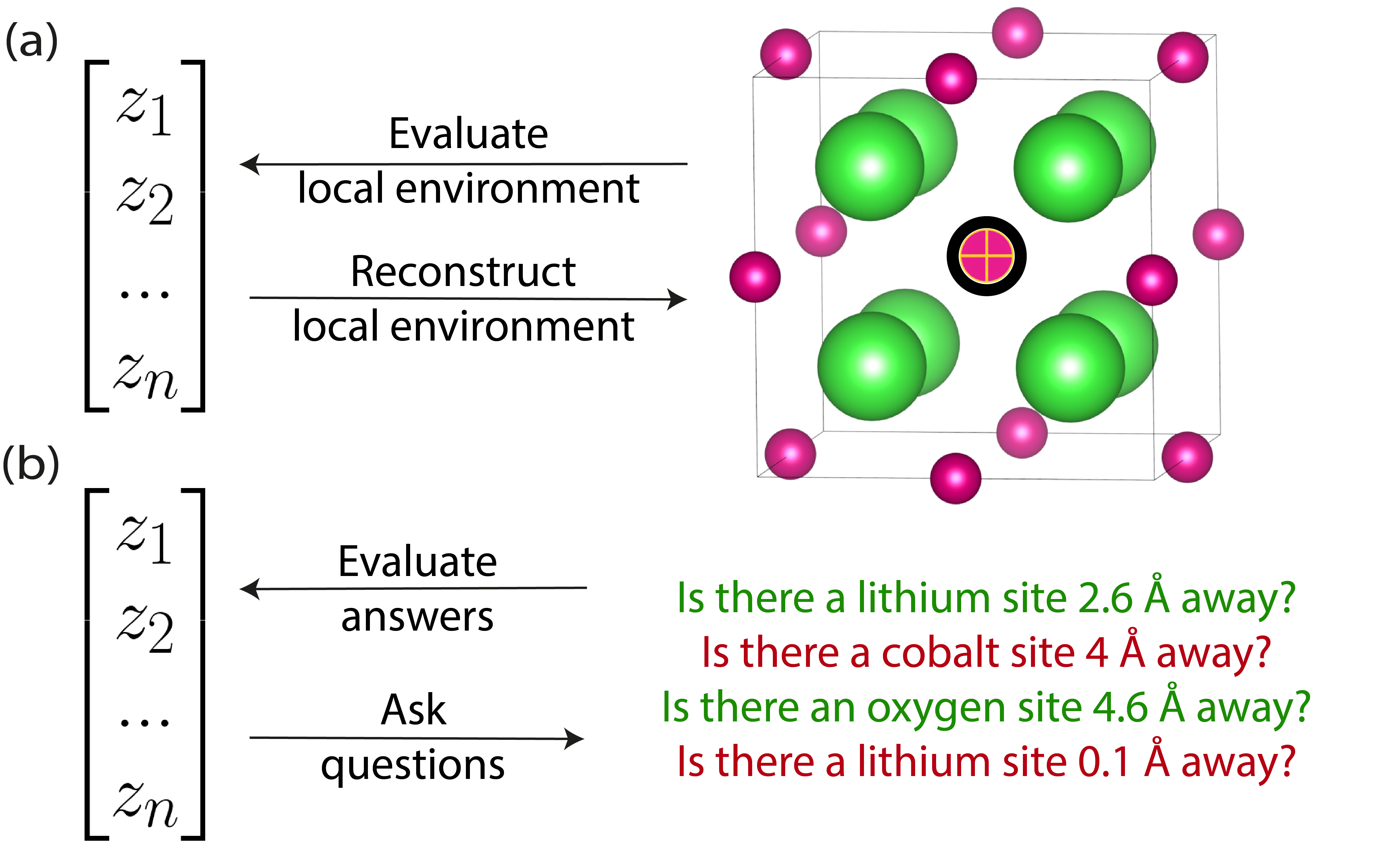}
	\caption[The intuition for how a Deep InfoMax model and an autoencoder evaluate the quality of a learned vector representation ($z$) is compared and contrasted.]{
     The intuition for how a Deep InfoMax model and an autoencoder evaluate the quality of a learned vector representation ($z$) is compared and contrasted.
     In each case, the considered representation is that of an oxygen local environment in the conventional unit cell of lithium oxide, the oxygen atom has been centred in the unit cell and highlighted.
     (a) In an  autoencoder, the model attempts to reconstruct the local environment from the latent representation; the model is then evaluated on the basis of the accuracy of the reconstruction.
     (b) Deep InfoMax avoids the requirement to reconstruct the local environment by answering an array of questions.
     The array of questions is generated by pairing the representation with both local samples used to create it and local samples from a different crystal; the questions are on which pairs belong to each other and which do not.
     In this case the samples are the individual neighbours within the local environment and their distance from the central atom, which are the constituents of the local environment set prior to pooling into a single vector.
     The questions are answered using a binary classifier is evaluated for its ability to answer the question of which neighbours belong to a local environment and which do not.
     The ability for the model to correctly identify all so called "true samples" and "false samples" for a representation verifies the information content of the representation in the same way that a successful reconstruction does in (a).
     More formally, the ability for the model to identify all constituents of the set given the representation acts as a lower bound on mutual information between the representation and the original input.     
	}
	\label{fig:Intuition_crystal}
\end{figure*}

Mutual information is the expectation of how much information can be gained about one random variable by knowing the value of another\cite{mutualinformation}; it is maximised when the variables contain identical information.
In the context of representing a crystal point set as a vector, mutual information between the crystal and the representation measures how much you can know about the crystal given only the representation, and vice versa, across the dataset.
Mutual information can be interpreted as one of many metrics for representation quality.
If the mutual information is very low, the representation does not contain much, if any, of the information present in the crystal which limits the usefulness of the representation.
If mutual information is maximised, then all of the information in the crystal is present in the representation and the parameters of the model are optimised to preserve information. 

As mutual information is an essential component of a good representation, it is an interesting target to explicitly maximise with self-supervised learning. Unfortunately, computing mutual information is notoriously difficult for complex data\cite{10.1162/089976603321780272, Walters-Williams2009, czyż2023normal} that cannot be fit with simple distributions, and even more difficult if we require the function to be differentiable so we can use it as a loss function for deep learning.
This is especially true in a self-supervised learning context since what often motivates self-supervised learning is that the input is complex and that our understanding is not sufficient to construct a fixed vector representation without heuristics, data driven or otherwise.
Fortunately, if the goal is to increase mutual information via a loss function, exact computation of mutual information is not necessary.
Instead, a gradient exposing estimate of mutual information that acts as a lower bound on mutual information is sufficient. If we maximise a lower bound, we are guaranteed to also increase the actual mutual information if it was initially below that lower bound.

Autoencoders, the most common kind of self-supervised learning architecture, can be understood in this mutual information context.
Autoencoders use an encoder to construct a representation and a decoder to reconstruct the input from the representation. Although autoencoder reconstruction loss is not an explicit formulation of mutual information, reconstruction loss can be shown to act as a lower bound on mutual information between the input and the representation produced by the encoder\cite{autoencoderMIproof}. If it is possible to always perfectly reconstruct the input from its laten representation, then mutual information is maximal.

Unfortunately, reconstructing the original crystal point set from a fixed vector representation is difficult because the point set representation of the crystal is discrete (a variably sized set) and the vector representation is continuous.
While it is trivial to reduce the point set to a continuous vector, the reverse is more challenging since it is unclear how to extract discrete qualities like how many atoms are there in the unit cell, and thus the size of the set.
Existing solutions to this discretization problem involve trade-offs and additional complexities.
Examples include a decoder that predicts crystal properties and then uses a separate diffusion network for reconstruction\cite{xie2022crystal}; sacrificing invariances on the output representation \cite{Alverson2022}; or using a latent representation that is not a vector\cite{tang2022graph}. As such, it would be desirable to be able to perform self-supervised learning on crystals, that can map the crystal to a vector, without having to solve the reconstruction problem and introduce additional complexities to the model.

Deep InfoMax is an alternative to the autoencoder (\cref{fig:Intuition_crystal}) that verifies the quality of the representations through a loss function that explicitly maximises the mutual information between the representation and the original input - without reconstruction.
This is performed with a classifier where the loss function of that classifier is an explicit lower bound on mutual information.
This classifier would receive a representation and a crystal, either the crystal that generated the representation or some other unrelated crystal, and would be tasked with determining whether that representation came from that crystal.
If the classifier can correctly identify whether a particular representation came from a particular crystal, and correctly identify when it did not, then the lower bound on mutual information between representation and input can be said to be maximised.
Since the output of this model is a scalar classification score, no additional complexity is introduced. This idea of verifying the quality of a representation using a classifier that queries the representation about the input used to generate it is the core intuition of Deep InfoMax.

\subsection{Maximising mutual information within constraints}

Any bijective mapping of input to representation will maximise mutual information, and any bijective transformation of that representation will preserve the encoded information.
For this reason, the naive maximisation of mutual information allows many pathological solutions.
These pathological encodings of the information are arbitrarily complex, and no realistic machine learning model would be able to make use of the encoding due to the heavy levels of obfuscation introduced by that complexity.
Consider taking a strong representation and then applying a reversible function that pseudorandomly swaps the positions of every input in the representation space.
This transformation would make the representation impossible to learn from by any real world model, but mutual information would none the less be preserved because there is some hypothetical function that would be able to reverse engineer the pseudorandom transformation and make full use of the information.

Simply maximising mutual information is insufficient; it is the usable information that matters, i.e, the information that a practical downstream machine learning architecture can effectively leverage.
This more specific concept of mutual information in terms of the information that is accessible to a particular model can be defined explicitly\cite{xu2020theory}, but directly computing this is not practical for use as a loss function.
Instead, by maximising mutual information within heuristically determined constraints, we can approximate this idea of maximising the "accessible mutual information", and achieve a stronger self-supervised learning pipeline.
The main component of our eventual loss function represents a naive maximisation of mutual information, the constraints limit the solution space to representations that a reasonable machine learning architecture can take advantage of.

The first architectural constraint is that the classifier that estimates the mutual information should be quite simple.
If the classifier is deep and has a lot of parameters, then there is no guarantee that the information exploited by the Deep InfoMax classifier will be readily accessible by a more simple downstream model and may be locked behind several complex, non-linear transformations of the representation.
In contrast, if a linear classifier can find large amounts of usable information in the representation, then it is known that the information measured by the linear classifier is also accessible to a downstream linear regression model.
Corollary to this, for the linear classifier, there is unlikely to be much value added for a non-linear model.
Ideally, the complexity of the classifier should be similar to the complexity of the downstream model.
In this work, the Deep InfoMax classifier uses a single hidden layer and a single non-linear activation function, which is aligned in complexity with the downstream supervised models used later in this work.

The second architectural constraint is to force the classifier to pay attention to the local properties of the input by only providing it with a local patch of the input instead of the full input.
Since we are using a set representation of the crystal, a local patch is just a constituent of the set.
For Site-Net, a local patch is either a single pairwise interaction between the centroid of a local environment and a neighbour, or a single local environment in the crystal.
We replace the question of whether the representation came from a particular crystal with an array of questions asking if each and every individual constituent of the crystal belongs to the crystal that produced the representation.
This approach forces the model to use the properties of the crystal to make decisions on what does and does not belong and to look at correlations between the properties of crystals across the dataset to assist in this.
By only allowing the model to see local patches of the input when making a judgement, it is less likely to overfit to the training data and more likely to cluster similar materials closer together.
This is the "local Deep InfoMax" objective as defined in the original implementation of Deep InfoMax\cite{hjelm2019learning} and the best performing models used this paradigm exclusively.
This makes graphs and sets ideal inputs for Deep InfoMax because what a "local patch" means is well defined, and querying over every characteristic of the representation makes sure that all of the information in the crystal is accounted for.
We exclusively use the local formulation of mutual information maximisation.

\begin{figure*}
	\centering
	\includegraphics[width=\textwidth]{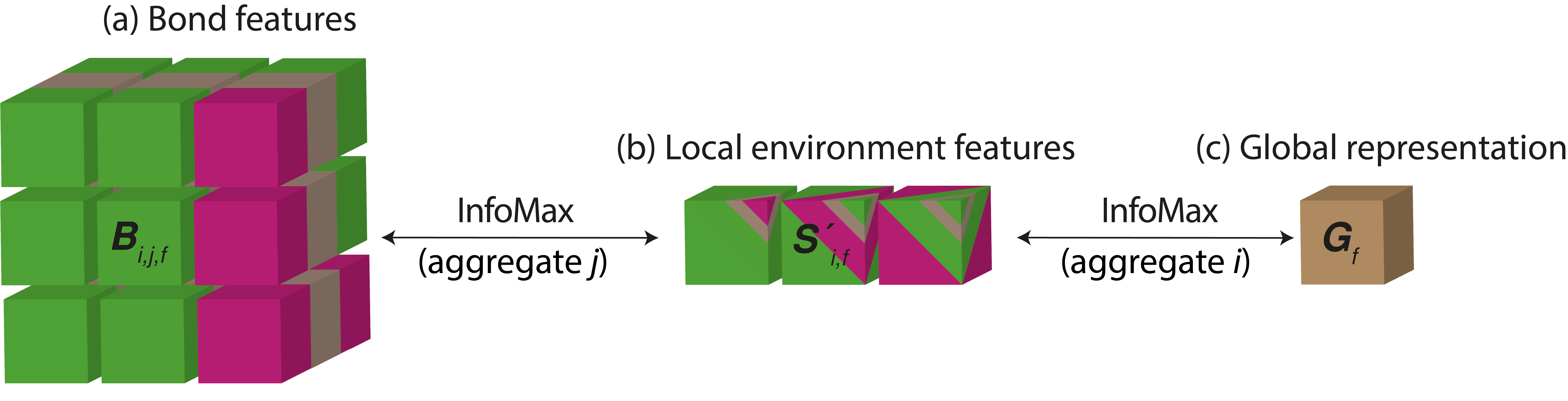}
	\caption[The Site-Net architecture represents the crystal as a (a, $B_{i,j,f}$) matrix of every bond in a large cubic supercell consisting of the elemental features of the sites, their neighbours, and the chemical interaction between them.]
 {The Site-Net architecture represents the crystal as a (a, $B_{i,j,f}$) matrix of every bond in a large cubic supercell consisting of the elemental features of the sites, their neighbours, and the chemical interaction between them. For demonstration purposes the features are shown for a primitive unit cell of Li$_2$O as per (\cref{fig:Intuition_crystal}). The colours in the bond features represent the information from the lithium features, the oxygen features, and the interaction between them which become increasingly mixed as the representation is reduced to a single vector. $i$ and $j$ represent which atomic site is being considered and are of length equal to the number of sites in the supercell. $f$ is the featurisation axis and varies in length throughout the model according to the size of the neural network layers and the starting feature vector length. Each row ($i$) and column ($j$) in $B_{i,j,f}$ can be understood to be a local environment centred on site $i$. Site-Net uses self-attention to pool pairwise interactions centred on a particular site into (b, $S^\prime_{i,f}$) local environment features, and the resulting set of local environment features are then pooled into a (c, $G_f$)  single global feature vector that describes the entire crystal. The colours demonstrate the increasing levels of information mixing from raw pairs of elemental features with a distance, to local environments, to a single feature vector summarising the crystal. The aggregation of pairwise interactions into local environment descriptors and the aggregation of local environment descriptors into a global feature vector represent distinct steps of aggregation where information loss is possible. Therefore, Deep InfoMax is applied independently to both parts of the process. Mutual information is maximised between the vector representing the local environment and its constituent pairwise interactions, and mutual information is maximised between the global crystal representation and its constituent local environments.}
\label{fig:MI_Steps}
\end{figure*}

The third architectural constraint is to introduce noise to the representation before providing it to the classifier.
By introducing noise at this stage of the model, the model is encouraged to place similar crystals closer together in the representation space because if it does not do this, then there is the risk that the noise will cause the classifier to incorrectly identify the crystal.
In our work, we add and regulate noise with the weighted Kullback–Leibler loss introduced in $\beta$-VAEs\cite{higgins2017betavae} as a regularisation term.
This term turns the magnitude of this noise into a trainable parameter of the model where the model tries to balance mutual information maximisation with maximising the tolerance to noise.

The fourth architectural constraint is to introduce additional "false samples", that is, the constituents that the Deep InfoMax classifier must identify as not belonging to the representation.
In addition to "false samples" taken from other crystals in the dataset, additional "false samples" can be generated that highlight different aspects of the chemistry of the material.
For an explicit lower bound on the information, it is necessary that false samples be uniformly taken from the rest of the data set without bias\cite{tschannen2020mutual}.
As such, this false sampling regime must always be included and encapsulated in its own loss term.
However, it does not need to be the only source of false samples contributing to the loss function, and additional false samples can be included to fine-tune which information is prioritised.
In practice, the performance of representations trained with Deep InfoMax on downstream tasks is strongly dependent on the false sampling, and performance can be improved dramatically by introducing additional false samples to the model that use domain knowledge to focus on the most difficult and relevant cases.
This is an intentional source of bias introduced to the model that encourages the model to focus on chemically relevant differences between samples and use chemical reasoning for constructing the latent space.

In summary, we introduce an alternative to autoencoders for self-supervised learning on crystals that does not require the crystal to be reconstructed by the model.
This is achieved by maximising the mutual information between the crystal and the learned representation using a classifier that determines which crystals and representations go together and which do not.
Since maximising mutual information on its own can lead to pathological solutions where the encoding of the information is too complex to be useful, we augment Deep InfoMax with a number of heuristic architectural constraints.
By maximising mutual information within these constraints the objective is to obtain an encoder that maximises the encoded information that is accessible to real world downstream models.

\FloatBarrier
\subsection{Independent learning of local environment and global representations}

Site-Net aggregates the crystal point set into a global representation vector in two steps using two independent permutation invariant aggregation functions.
The first step aggregates each sites interactions with all other sites in the crystal into a vector summarising the local environments.
The second step aggregates the local environments into a global feature vector summarising the entire crystal.
We apply Deep InfoMax at both stages of aggregation and treat each aggregation as a separate self-supervised learning pipeline.
For learning the local environments, Deep InfoMax takes a learned representation of a local environment produced by the attention block, a pairwise interaction from that environment, and a pairwise interaction from an unrelated local environment.
The model determines which pairwise interaction belongs to the local environment and which does not.
To learn a representation of the entire crystal structure, Deep InfoMax takes a learned representation of the crystal, a local environment from that crystal, and a local environment from a different crystal, and determines which local environment belongs to the structure and which one does not.
This two-step process ensures that all the information about the crystal is available to the global representation (\cref{fig:MI_Steps}) where each step aggregates a properly defined set with no ordering information.
If we tried to maximise mutual information between a global crystal representation and pairwise interactions, we would lose the information about which pairwise interaction belongs to which particular local environment.

\begin{figure*}[]
	\centering
	\includegraphics[width=\textwidth]{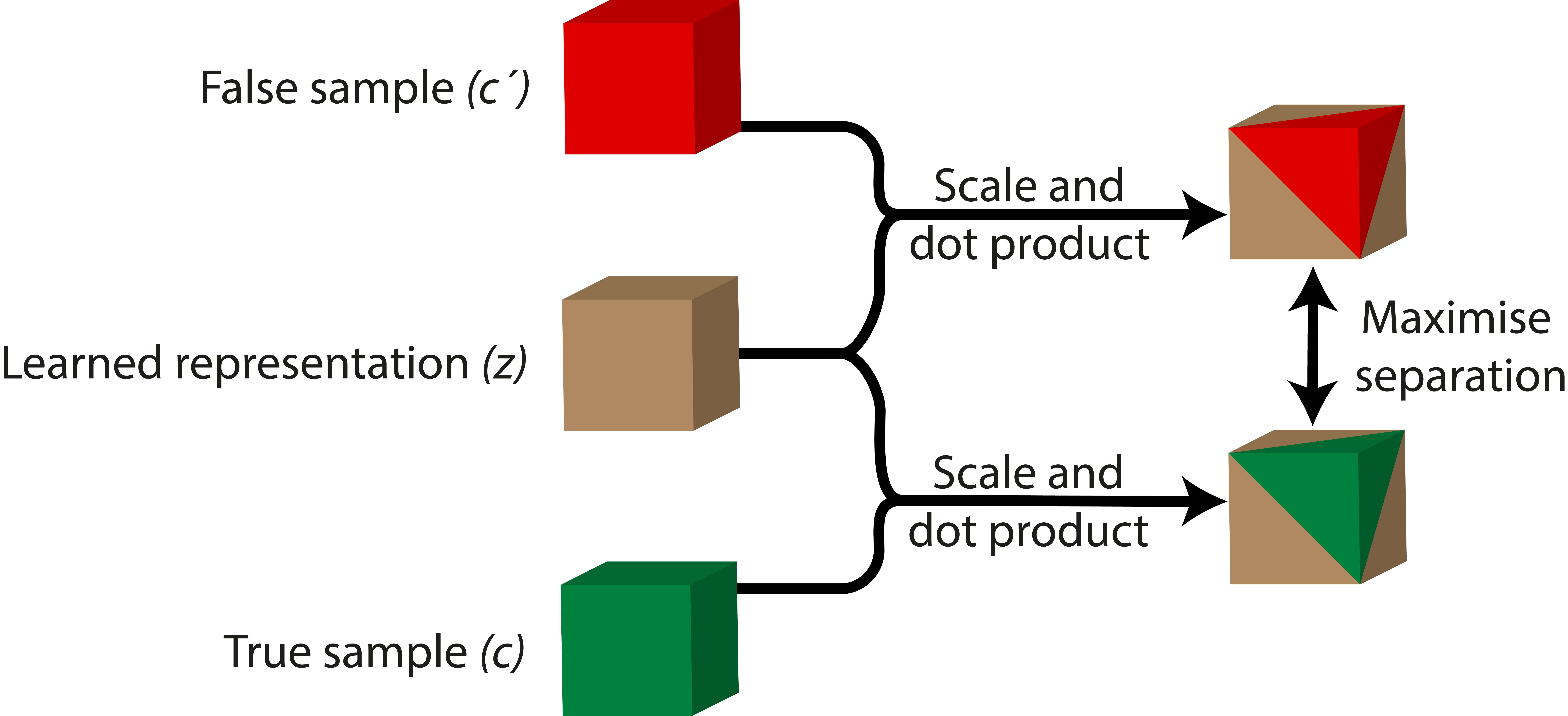}
	\caption[The computation of the Jensen-Shannon entropy loss function is visualised.]{The computation of the Jensen-Shannon entropy loss function is visualised. The representation ($z$), true sample ($c$) and false sample ($c^\prime$) are first upscaled to a shared higher dimensional space. Once in this higher dimensional space, the dot product is taken between the learned representation and the two samples. The Jensen-Shannon entropy loss function is then used to maximise the separation between the two dot products. The greater this separation across the dataset, the greater the lower bound on mutual information between the representation and the constituents from which it was created.
	}
	\label{fig:DIM_Loss}
\end{figure*}

\subsection{The Deep InfoMax Loss Function} 

As discussed above, we create an encoder that reduces the constituents of the crystal point set ($c$) to vector representations in a shared basis ($z$).
We then use a classifier to verify whether it was possible to identify which representations map to which crystal.
It is essential that this classifier acts as a lower bound on mutual information so that we have guarantees on the value of the mutual information for a given loss.
We now formally define the loss function used in Deep InfoMax. As Deep InfoMax is applied for self-supervised learning at both the local environment and global crystal representation level, we maintain the use of the generic set notation to describe the construction of the loss function.
In the former case, the constituents are the pairwise interactions and the learned representation vector is of the local environment.
In the latter case the representation vector is of the entire crystal and the constituents are the local environments produced by the previous part of the model.

\begin{equation}
\centering
J^s(z,c,c^\prime) = S^p(u^z(z) \cdot u^c(c)) + S^p(-u^z(z) \cdot u^c(c^\prime)) \quad\textrm{where}\quad S^p(x) = ln(1+e^x)
\label{eq:Js}
\end{equation}

The core loss function, the classifier used to estimate and maximise mutual information (\cref{eq:Js}), is the Jensen-Shannon entropy ($J^s$).
This classification loss quantifies the ability to determine whether a particular constituent belongs to a particular representation.
The Jensen-Shannon loss accepts a representation ($z$), a constituent of the set used to construct the representation ($c$), and a false sample from a different set that the representation was not constructed from ($c^\prime$). To place a lower bound on mutual information, the false sample is uniformly sampled from every available false constituent in the training data without bias. In practice, false constituents are randomly chosen from the same training batch. If the representation ($z$) is the vector featurising the entire crystal, then the constituents ($c$) are the local environment feature vectors it was created from. If the representation ($z$) is a local environment, the constituents ($c$) are pairs of atoms.

To compute the loss (\cref{fig:DIM_Loss}), we first scale the representation and samples to a shared higher dimensional space using a feed forward neural network, the size of which is a hyperparameter of the model. The model utilises two separate neural networks to perform this scaling: one dedicated to scaling the representation to a higher dimensional space ($u^z$) and another for scaling the samples ($u^c$).

Once scaled, we compute the dot product between the upscaled representation and the upscaled samples, resulting in a true score for the representation true sample pairing and a false score for the representation false sample pairing. Summing the softplus ($S^p$) of these two values is the Jensen-Shannon entropy, and as the Jensen-Shannon entropy gets lower, the lower bound on mutual information between the representation and its constituents is increased.

It is noted that the Jensen-Shannon entropy could be replaced with any classifier, and the same result would be achieved of teaching the model to distinguish true and false samples. The Jensen-Shannon entropy has a number of mathematical qualities that make it a pragmatic loss function for our approach, in addition to acting as a lower bound on mutual information. The first is that it is a smooth and monotonic loss function, and the second is that there are diminishing returns for correct answers, so the model does not focus exclusively on easy cases.

Once again, it is noted that maximising mutual information is not sufficient on its own for constructing a representation, as a representation space could have very high mutual information with the crystal but have an unusual shape that makes it difficult to interpret for all but the most complex models. Maximising mutual information using local patches of the input does a lot of work in forbidding trivial solutions, but they are still possible. To mitigate this possibility further, a regularisation term assists in learning the representation by forcing it to conform to a prior distribution that is known to be well behaved and easy to work with, such as a uniform or gaussian distribution. To this end, in addition to just the Jensen-Shannon loss, we incorporate a regularisation term that sculpts the representation space.

\begin{equation}
\centering
J^s(z+\sigma y,c,c^\prime) = S^p(u^z(z+\sigma y) \cdot u^c(c)) + S^p(-u^z(z+\sigma y) \cdot u^c(c^\prime))
\label{eq:jsperturb}
\end{equation}

\begin{equation}
\centering
K^l(z,\sigma) = mean(z^2 + \sigma - ln(\sigma))
\label{eq:kl_divergence}
\end{equation}

\begin{figure*}
	\centering
	\includegraphics[width=\textwidth]{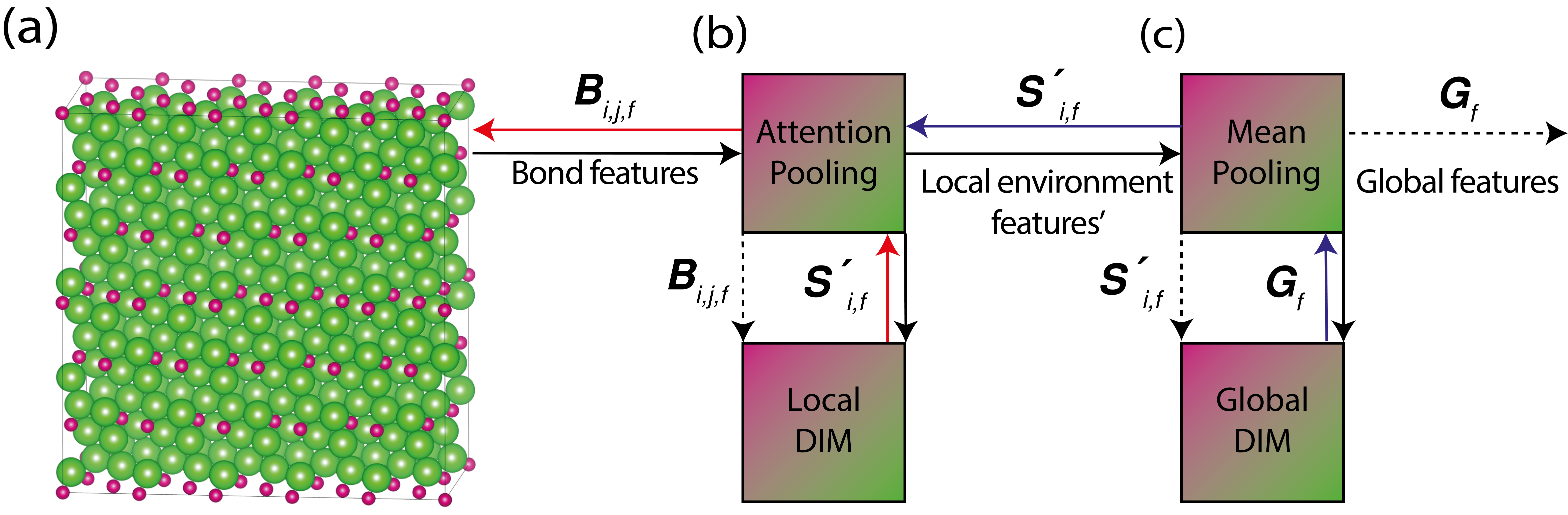}
	\caption[The learning of a global representation of the crystal takes place in two distinct steps.]{
	The learning of a global representation of the crystal takes place in two distinct steps. The first step is to generate summaries of the local environments of the sites in the crystal using a Site-Net transformer block and to maximise the mutual information between local environment features and the pairwise interactions they are constructed from. The second step is to process the local environment features through some shared neural network layers before taking the mean to construct a global feature vector. Mutual information is maximised between the learned local environment features and the global feature vector. These are effectively two separate models, and the gradients are isolated from each other. The global Deep InfoMax objective cannot adjust the features of the local environments. The backpropagation is shown explicitly using coloured arrows. The gradient flow for the local environment feature learning is shown in red, and the gradient flow for the global feature learning is shown in blue. Dashed lines represent the forward propagation of features without corresponding back propagation. The two essentially independent sub models are trained in parallel.
	}
	\label{fig:DIM_flow}
\end{figure*}

There are two components to the regularisation. The first is that before introducing the representation to the classifier (\cref{eq:jsperturb}) we perturb it with noise by adding a vector sampled from a unit Gaussian of the same dimensionality as the representation ($y$) multiplied by a learnable vector of scalars ($\sigma$) that control the magnitude of the noise. Note that this is the same "Reparameterisation Trick" used in variational autoencoders\cite{Kingma_2019}, though for the sake of accessibility we do not adopt the variational autoencoder formalism for describing its impact on the model. The purpose of this noise is to ensure smoothness under local perturbations in the latent space by clustering similar crystals together. The reason this encourages clustering of similar crystals is that the classifier can receive any feature vector in the local region around the representation once the noise is applied. If nearby crystals are similar, it is more likely that the classification is correct regardless. The second is that we control the magnitude of the representation vectors. Without an explicit penalty for making the representation vectors large, the noise can be trivially bypassed by inflating the magnitude of the latent space such that the noise is arbitrarily small in comparison. 

The Kullback–Leibler divergence ($K^l$) term is the regularization loss term that combines the noise component and the magnitude component (\cref{eq:kl_divergence}). Formally, it measures the distance between the distribution defined by the values of the representation after applying noise, and a unit Gaussian. In terms of its function for training, the Kullback-Leibler divergence term rewards the model for maximising the ratio of noise ($\sigma$) to magnitude ($z^2$) for a given lower bound on mutual information. The reason for this is that it is assumed that if the same degree of mutual information can be achieved with more noise, then the latent space with greater noise tolerance is more useful as a representation.The Kullback-Leibler divergence term is computed for each dimension in the representation vector, and then the mean value is taken. The mean is taken rather than the sum so that the magnitude of the loss is independent of the dimensionality of the representation.

\begin{equation}
\centering
\text{Loss} = \sum\limits_{i=1}^{I}\frac{\alpha}{I} J^s(z+\sigma y,c_i,c^\prime_i) + \beta K^l(z,\sigma)
\end{equation}

The final loss is a weighted sum of the Jensen-Shannon entropy (\cref{eq:Js}) and the Kullback–Leibler divergence (\cref{eq:kl_divergence}), where $\alpha$ and $\beta$ are hyper parameters. The ratio of the parameters $\alpha$ and $\beta$ will determine the balance between a highly regularised representation, and a high mutual information representation. Optimal performance is expected from a careful balance of the two terms. We find that an $\alpha$ of 1 and a $\beta$ of 0.1 give good results and outperform the zero regularisation case, but no further explicit optimisation has been performed on the $\alpha$ and $\beta$ ratio. The Jensen-Shannon entropy's contribution is averaged over the number of constituents in the set ($I$), be it the number of atoms in the local environment or the number of local environments in the unit cell. The Kullback-Leibler divergence is normalised with respect to the dimensionality of the representation vector. The purpose of these normalisations is to prevent the size of the crystal and the number of dimensions in the representation vector from implicitly altering the relative magnitude of the loss components.

\subsection{Implementation}

We construct a Site-Net with a single head and a single attention block and with two distinct Deep InfoMax loss functions that are respectively used to optimise the representation of the local environments and then the global representation (\cref{fig:DIM_flow}). The attention block aggregates the pairwise interactions that constitute a local environment into local environment feature vectors, and the local Deep InfoMax loss function is used to maximise mutual information between these local environment features and their constituent pairwise interactions. These learned summaries of the local environments are then passed through shared neural network layers before being mean pooled into global representations of the entire crystal; the global Deep InfoMax loss function is then used to maximise mutual information between the global aggregates and the local environments from which they were constructed.

\begin{table*}[t]
    \caption[The hyperparameters for the Site-Net trained with Deep InfoMax are shown.]{The hyperparameters for the Site-Net trained with Deep InfoMax are shown. These hyper parameters are a reduced version of the hyper parameters found by the hyper parameter search in the original Site-Net work. All hyper parameters have the same meaning as per the original Site-Net implementation with the exception of the upscaling layers used as part of the Deep InfoMax classifier. It was also found that including the coulomb force as an interaction feature suppressed performance, so only the Euclidean distance matrix was used.
    } 
    \centering
    \renewcommand{\arraystretch}{1.5}
    \begin{tabular}{p{0.4\textwidth}p{0.3\textwidth}p{0.2\textwidth}}
    \hline
    \hline
        Hyperparameter & Value used\\ \hline
        site features (from Pymatgen \cite{ONG2013314} \& Matminer \cite{WARD201860}) & 9: atomic number, atomic weight, row, column, first ionization energy, electronegativity, atomic radius, density, oxidation state\\ 
        site features length & 64\\ 
        interaction features (from Pymatgen \cite{ONG2013314}) & 1: distance matrix \\ 
        interaction features length & 64\\ 
        attention blocks & 1\\ 
        attention heads & 1 \\ 
        attention weights network [layers] & [64] \\ 
        upscaling network ($u^s$ and $u^p$) [layers] & [64, 128] \\ 
        pre-pooling network [layers] & [64, 128]  \\ 
        post-pooling network [layers] & [64]\\ 
        activation function & mish \cite{mish} \\ 
        optimizer & adamw\\ 
        learning rate & 8.12$\times$10$^{-4}$ \\ 
        normalization method & layernorm \cite{layernorm} \\ 
        global pooling function & mean \\ 
        batch size (unique sites) & 1200 unique sites (400 for 100 sample supervised learning) \\
        \hline \hline
    \end{tabular}
    \label{Tab:hparam}
\end{table*}

The hyper parameters of the model were chosen by downsizing the hyper parameters reported in the original Site-Net (\cref{Tab:hparam}) both for computational speed and to achieve a lower dimensional representation for use with small supervised datasets. The size of the supercells was also reduced to 50 from 500 and materials with more than 50 atoms in the primitive unit cell were excluded from the data sets. This reduced the size of the training dataset for the band gap from 84890 samples to 70590, and the size of the formation energy dataset from 106201 samples to 88740, but made models significantly faster to train. This weakens the assumption of the supercell being cubic that Site-Net relies on and as such weakens the performance but this does not impact comparisons between models. The final hyper parameters are heuristic and were chosen to achieve reasonable model performance with minimal computational overhead.

\subsection{Synthetic Samples}
As discussed previously, a key part of optimising the Deep InfoMax model is the selection of false samples provided to the classifier. The methodology in the original implementation of Deep InfoMax pulls false samples from other crystals in the dataset, which is a requirement for the classifier to act as a lower bound on mutual information. The exclusive use of these false samples has limitations when trying to learn a representation of the crystal structure. Specifically, in almost all cases, it is trivial to discriminate true samples from false samples solely on the basis of composition, unless the materials are polymorphs. For example, no knowledge of the structure is required to conclude that there is no carbon in a NaCl local environment. The vast majority of the variance in crystal databases can be explained through stoichiometry alone, so there is little incentive to encode structural information without some bias in the training.

\begin{figure*}[!ht]
	\centering
	\includegraphics[width=\textwidth]{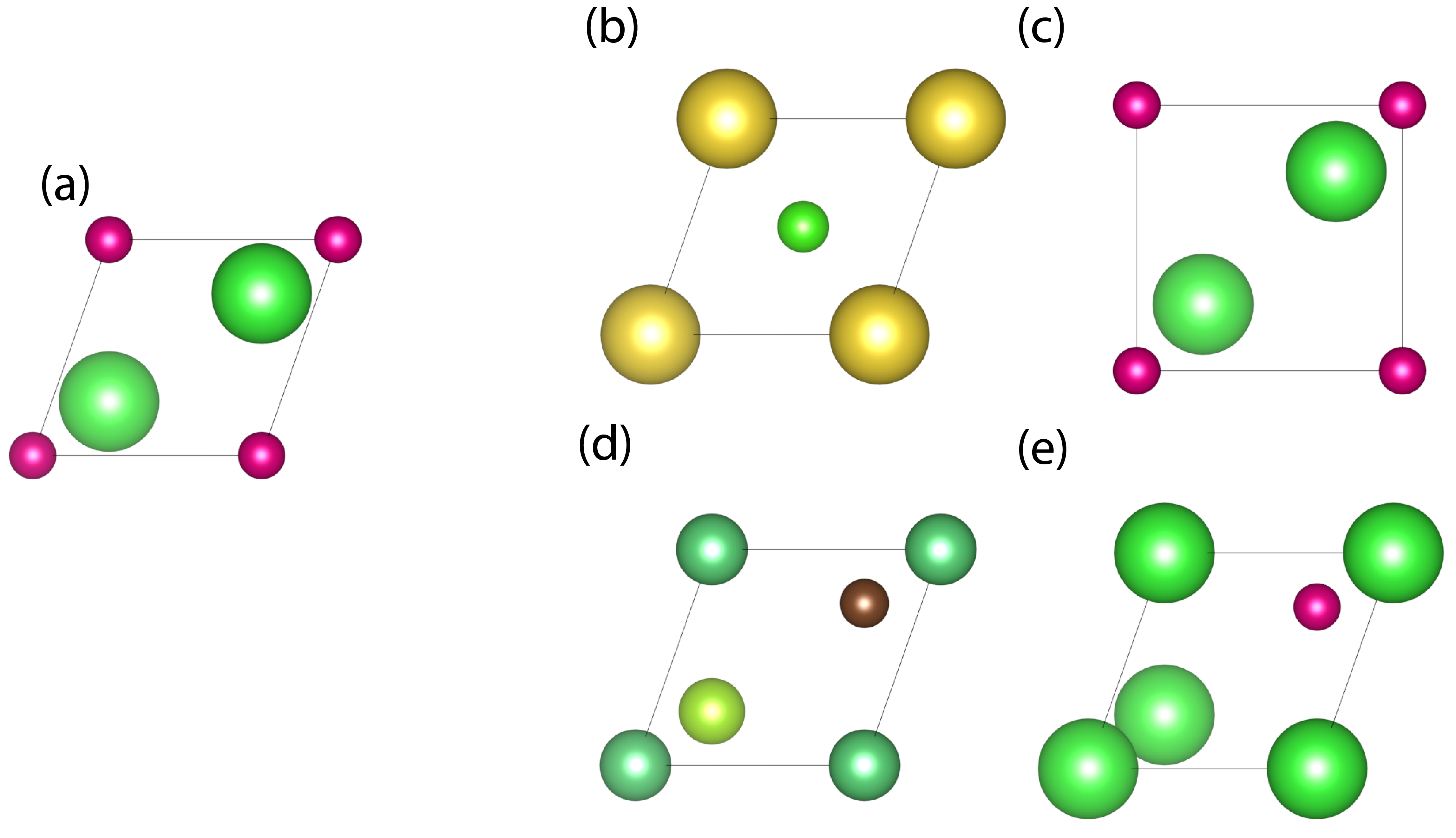}
	\caption[]{
	The false sampling strategy is shown in context of (a) Li$_2$O. The false samples consist of (b) an unrelated real crystal, (c) an artificial crystal with the same stoichiometry as Li$_2$O with geometry donated from an unrelated real crystal, (d) an artificial crystal with the same structure as Li$_2$O but with stoichiometry donated from an unrelated real crystal, and (e) the same stoichiometry and structure as Li$_2$O but the positions of each atomic species are randomised. The final false sample with shared structure and stoichiometry is only deployed when learning global representations from the local environment representations, as the risk of "sample collision", that is, false samples being the same as true samples by chance, is too high when considering only two elements and a distance.
	}
	\label{fig:False_Samples}
\end{figure*}

To alleviate this issue, we engineer synthetic false samples that cannot be discriminated on composition alone, such that we can bias the representation towards structural information. Notably, these synthetic false samples do not need to be stable or even realistic. What matters is that it should be difficult to discriminate these synthetic false samples from the true sample without using structural information.

Three different kinds of synthetic false samples are used in addition to sampling other crystals directly (\cref{fig:False_Samples}). The first kind of synthetic false sample is the "false polymorph" generated by taking the structure from another crystal in the dataset but keeping the site identities of the true crystal. This false polymorph will have an identical composition, so the change in structure must be relied upon to identify the fake. The second kind of false sample is where the structure is identical but the composition is taken from another crystal, changing the elements present at each site. This false sample is introduced so that the model can learn compositional information independently of the positioning of the sites. The final kind of false sample is the most difficult for the model; it is a "false permutation" where the structure is the same and the composition is the same, but the site positions are permuted. For the false permutation, the classifier must combine the structural and composition information, neither of which contains the information on its own. For the false permutation, there are always sites where there should be sites, and the composition is correct, and the model must understand when the site at a particular position is of the wrong species to catch the fake. The false permutations are only used in the second step of aggregation, if the false permutations are employed when learning local environment features then any permutations between sites of the same element will result in those neighbors remaining correct.

\begin{equation}
\centering
\text{Loss} = \sum\limits_{i=1}^{I}\sum\limits_{j=1}^{J}\frac{\alpha}{IJ} J^s(z+\sigma y,c_i,c^\prime_{ij}) + \beta K^l(z,\sigma)
\label{eq:mod_js}
\end{equation}

In computing the Jensen-Shannon entropy, the new false samples are introduced in parallel to the false samples randomly sampled from other real crystals. To modify the loss function to account for multiple false samples (\cref{eq:mod_js}), a false sample is provided from all sources ($c^\prime_{ij}$). The mean Jensen-Shannon entropy for all false samples is taken.

Optimising the false-sampling regime with domain knowledge will likely lead to significant improvements in the quality of the representation moving forward. In this work, the engineered false samples were limited to those that could be generated during training by moving data between crystals, but more complex false samples could be generated before training using physical processes. Equally, additional true samples can be generated in a similar manner. For example, small thermal perturbations to the positions of each atom in the unit cell structure could be used as adjunctive true samples to teach the model not to consider thermal noise as changing the crystal's identity. Combining more complex sample generation with adjunctive true samples, one could consider defect chemistry or assert that adding thermal noise to true samples does not change their identity.

Given the nature of the sampling, it is possible that a false sample almost or exactly matches a true sample from the material. This puts a limit on the degree of confidence in the classification of samples that are quite common. The effect of sample collisions and to what extent they should be explicitly prevented is a matter for future investigation. Collisions are significantly more likely with the local Deep InfoMax loss, since the considered pairwise interactions are just two atoms and a distance, while the local environment features only collide if all of the pairwise interactions for two environments are identical. Site-Nets use of supercells makes samples collisions significantly less likely when learning the global representation.

\begin{figure*}[!ht]
	\centering
	\includegraphics[width=\textwidth]{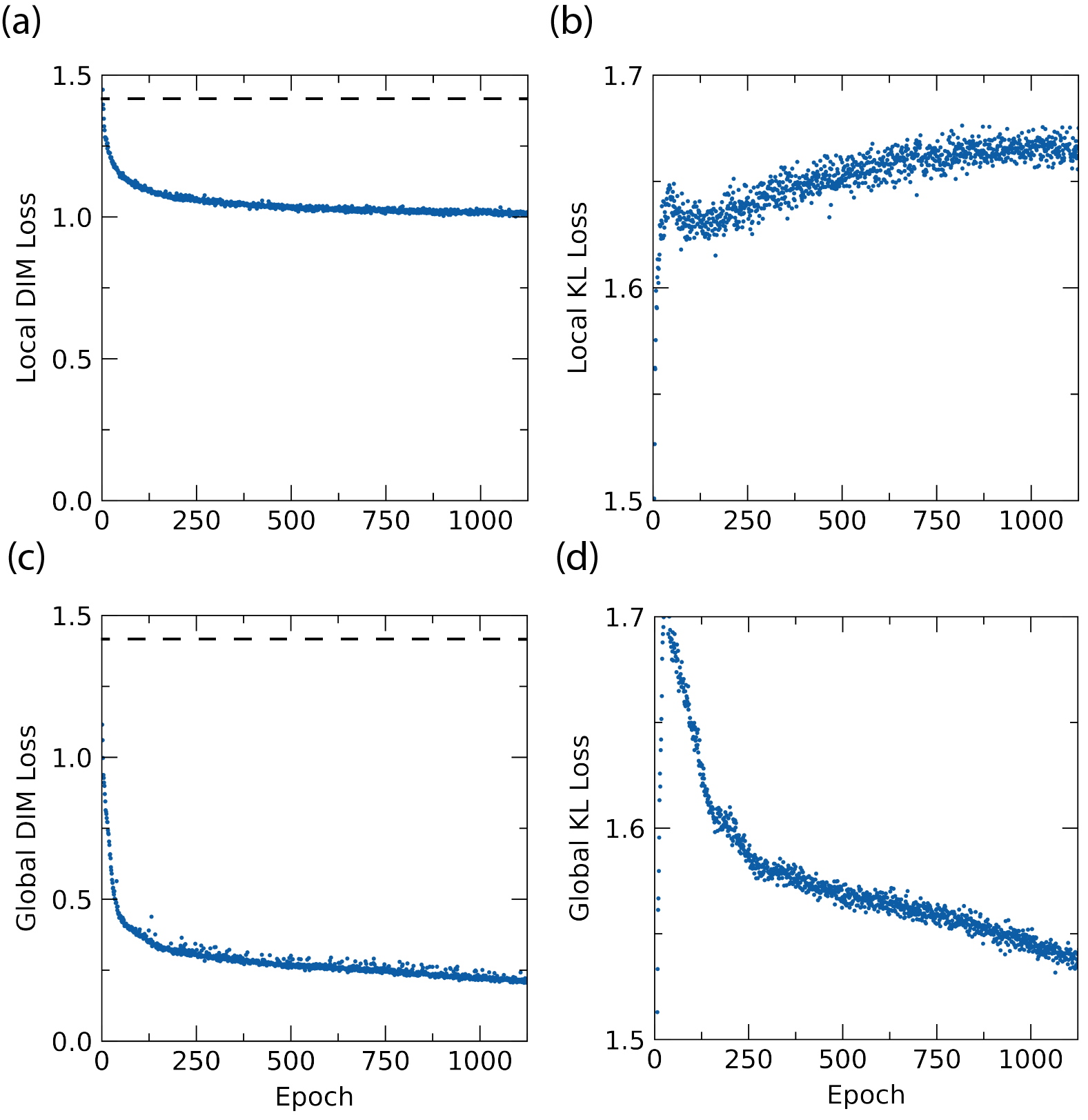}
	\caption[]{
        The validation learning curves are shown for a Deep InfoMax model trained on the formation energy dataset, where for all loss functions a lower value is better. Shown are the validation curves of the (\cref{eq:Js},a,c) Deep InfoMax (DIM) loss and the (\cref{eq:kl_divergence},b,d) KL divergence loss, for both the (a,b) local and (c,d) global encoders, plotted against the training epoch. The DIM loss is averaged over all false samples as per \cref{eq:mod_js}. The Deep InfoMax training is fairly smooth and monotonic for both encoders, with the global encoder generally being more successful in maximising mutual information. To give context to the Deep InfoMax loss values, the baseline loss corresponding to always assigning a classification score of zero is shown with a dashed red line. The KL divergence loss acts as a constraint on the model. There is an upward surge in the KL loss early in training where the model must violate the constraint to make performance gains on the Deep InfoMax, followed by an unstable equilibrium between the Deep InfoMax and KL divergence loss where the model makes trade-offs between the two to minimise the overall loss. The KL divergence loss is a stronger constraint on the local encoder than on the global encoder and remains high throughout training.
	}
	\label{fig:training plot}
\end{figure*}

\section{Results}

A key use case for Deep InfoMax in materials science is the potential to improve property prediction performance when only a small number of property labels are available. Large databases of crystal structures without specific property labels can be leveraged with Deep InfoMax. The intrinsic information about crystals in these large databases can then be used to improve the performance of the property prediction task. We investigate this use case for Deep InfoMax using the formation energy and band gap tasks in the Matbench\cite{Dunn2020} suite, the largest property prediction tasks available. These datasets contain crystal structures from the Materials Project\cite{10.1063/1.4812323} stored as Pymatgen structure objects along with their DFT band gaps and DFT formation energies and contain $\sim 10^5$ crystals along with their respective property labels.

\subsection{Data processing and Task Definitions}

To simulate a low data environment, we mask the majority of the labels in the band gap and formation energy datasets. The masked data can be used for Deep InfoMax pretraining, but not for property prediction, giving us a large unlabelled dataset for pretraining and a labelled subset of the data for supervised training. From there it can be determined if pretraining with Deep InfoMax improves performance on the smaller amounts of labelled data. This masking approach is chosen because it removes the distributional shift between the unlabelled dataset and the labelled dataset as a performance factor, since they are now from the same source. This is important because if an external dataset were used for Deep InfoMax pretraining, it would be difficult to distinguish the effects of Deep InfoMax from the effects of the chosen dataset. The two largest tasks in the Matbench suite are used because it maximises the amount of property labels that can be used while making sure that the crystals without property labels greatly outnumber the labelled crystals. If we mask all but 1000 property labels, there are 100 times as many crystals used in the pretraining as are made available for supervised learning.

To handle the data, we first split the datasets into train and test data in a 80/20 split, using the first fold as defined by Matbench to train the Deep InfoMax model. The 20\% of the data allocated to the test dataset is used to evaluate the performance of supervised models and is not used as part of supervised or Deep InfoMax training. The training characteristics of the Deep InfoMax (\cref{fig:training plot}) model are shown, in terms of the Jensen-Shannon entropy and the Kullback–Leibler divergence as a function of epoch.

To perform supervised learning, the property label masking is then applied to the training fold, with the removal of property labels being randomised between supervised models. This gives us a large collection of unlabelled crystals for Deep InfoMax pretraining and a smaller labelled dataset for property prediction. We test Deep InfoMax at a variety of property label availabilities. We work with 50, 100, 250, and 1000 available property labels.

We investigate two methodologies for applying Deep InfoMax. The first methodology is transfer learning, where the parameters from the Deep InfoMax model are used as the starting parameters for a supervised Site-Net model with the same architecture except for appending two additional neural network layers to map the produced global feature vector to a property prediction. These final layers are initialised randomly. The second methodology is representation learning, where we take the fixed vector representation of the crystal that the Deep InfoMax model produces and use it to train simple supervised models in scikit-learn\cite{sklearn}. These simple models consist of a linear regressor and a neural network with a single 64 node hidden layer. In this paradigm, the trained Deep InfoMax model is treated as a featuriser, a function that transforms a crystal structure into a vector.

\begin{figure*}[!ht]
	\centering
	\includegraphics[width=1\textwidth]{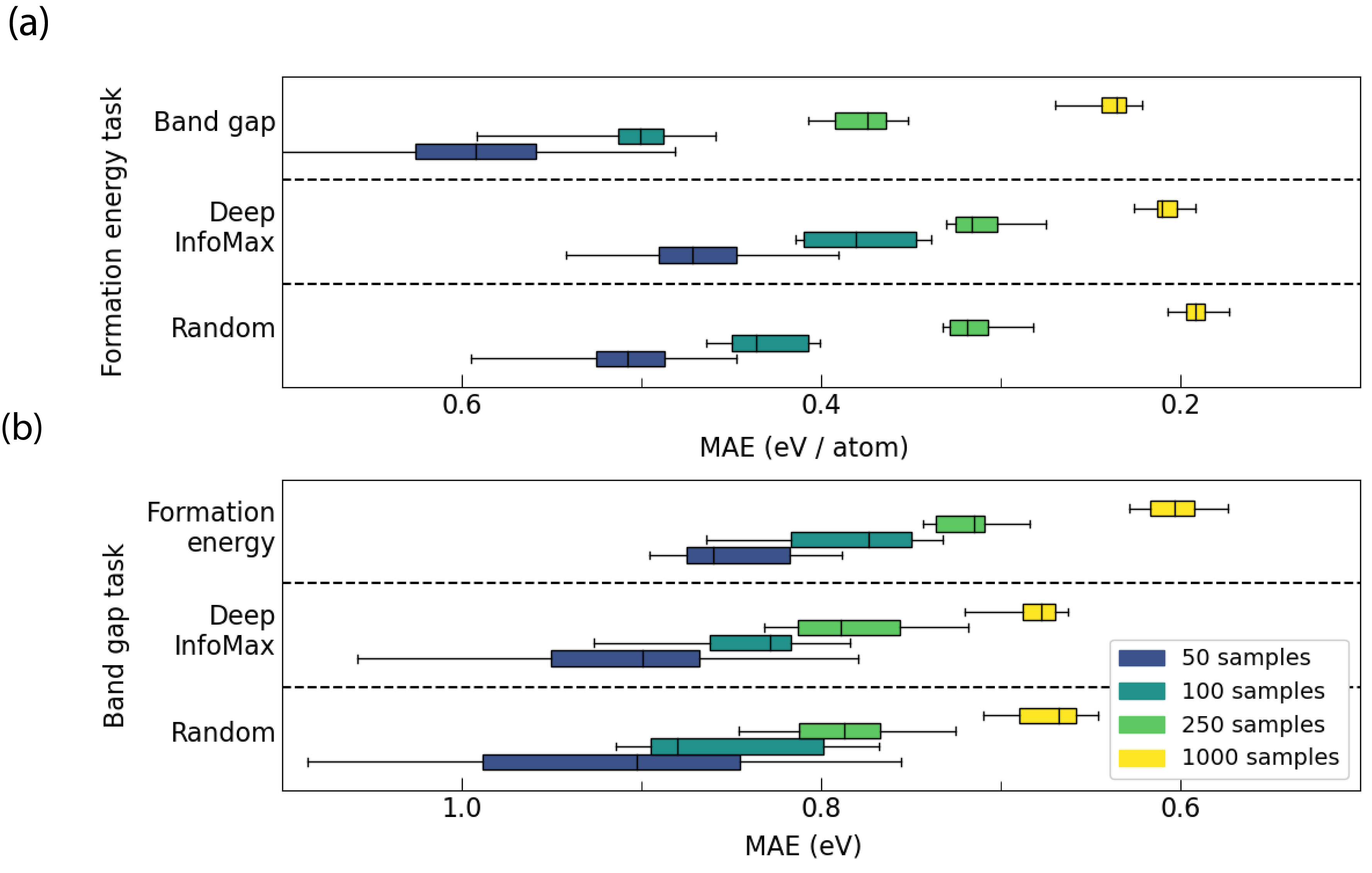}
	\caption[Box and whisker plots for the test dataset MAE are shown for supervised Site-Nets trained on the (a) formation energy and (b) band gap tasks, as a function of the amount of training data, and the starting parameters of the model.]{Box and whisker plots for the test dataset MAE are shown for supervised Site-Nets trained on the (a) formation energy and (b) band gap tasks, as a function of the amount of training data (50, 100, 250, 1000), and the source of the starting parameters of the model (random, Deep InfoMax, band gap, formation energy). Each box and whisker plot represents 12 supervised Site-Net models for the relevant starting parameters trained on a different random sub sample of the training data as determined by 12 shared random seeds. We compare supervised Site-Nets that started from random starting parameters, the parameters of a trained Deep InfoMax model trained on the full training data, and the parameters from a supervised Site-Net model trained on all available training data from the other task. This allows us to compare Deep InfoMax pretraining with supervised transfer learning and a random control. We demonstrate in the case of formation energy that Deep InfoMax pretraining improves the performance of downstream models with small amounts of data and is stronger than transfer learning from a band gap model. In the case of band gap, transfer learning from the formation energy task achieves the best performance, and there are small gains in performance with smaller amounts of data. For both tasks, transfer learning from Deep InfoMax does not improve performance with larger amounts of data.
	}
	\label{fig:transfer}
\end{figure*}

\subsection{Transfer Learning}

For deep learning, the parameters on which a model converges are dependent on the starting parameters. If the starting parameters are particularly poor, then the model may diverge and fail to train. If the starting parameters are particularly good, then the model can converge faster and to a better solution. Normally, the starting parameters of a model are set to random values using some distribution.

An alternative to random values is transfer learning, where the parameters of a donor model are used as the starting point for another model and then fine-tuned to the new task. Generally, the donor model was trained with more data, and there is reason to believe that the auxiliary information from that task can assist the model with access to only a small amount of data. As Deep InfoMax is a self-supervised task that works with the intrinsic information in the crystal, it is expected that this should correlate well with most properties. Alternatively, a model trained on another supervised task can be used for transfer learning.

To investigate the effects of transfer learning from Deep InfoMax on supervised Site-Net models, we train a Deep InfoMax model on all of the training data for the band gap and formation energy tasks. We then add an additional layer (initialised with random parameters) to the architecture that maps the output global feature vector to a property prediction and remove the Deep InfoMax classifiers. This modifies the network into a supervised Site-Net model. If we initialise this supervised model with random parameters before training, it is a standard supervised Site-Net model; if we initialise it with the parameters arrived at by Deep InfoMax training, then we are performing transfer learning from Deep InfoMax to the supervised task. Since we are working with two property prediction tasks, we can also use supervised transfer learning as a baseline by training supervised Site-Nets on the full training fold for each dataset and using these models as starting points for the other dataset. We train supervised Site-Nets initialised with random parameters, parameters taken from the Deep InfoMax model, and parameters taken from a supervised model training on the entire training dataset from the other task. We train 12 supervised Site-Nets in each category, and the nth trial for each model uses the same random seed for masking the training data to ensure direct comparability. Each supervised Site-Net is evaluated according to its MAE on the test dataset containing $\sim$20,000 samples with $\sim$20,000 property labels.

From the results (\cref{fig:transfer}), it is shown that transfer learning using Deep InfoMax parameters results in improvements in performance across both tasks in the lower data regimes. With greater amounts of labels available for supervised training ($>250$) the Deep InfoMax pretraining is harmful to performance. In general, Deep InfoMax pretraining is shown to be more effective for the formation energy task than the band gap task. In the case of the band gap task, supervised transfer learning from formation energy is uniformly the best option for all levels of label availability.

\begin{figure*}[!ht]
	\centering
	\includegraphics[width=1\textwidth]{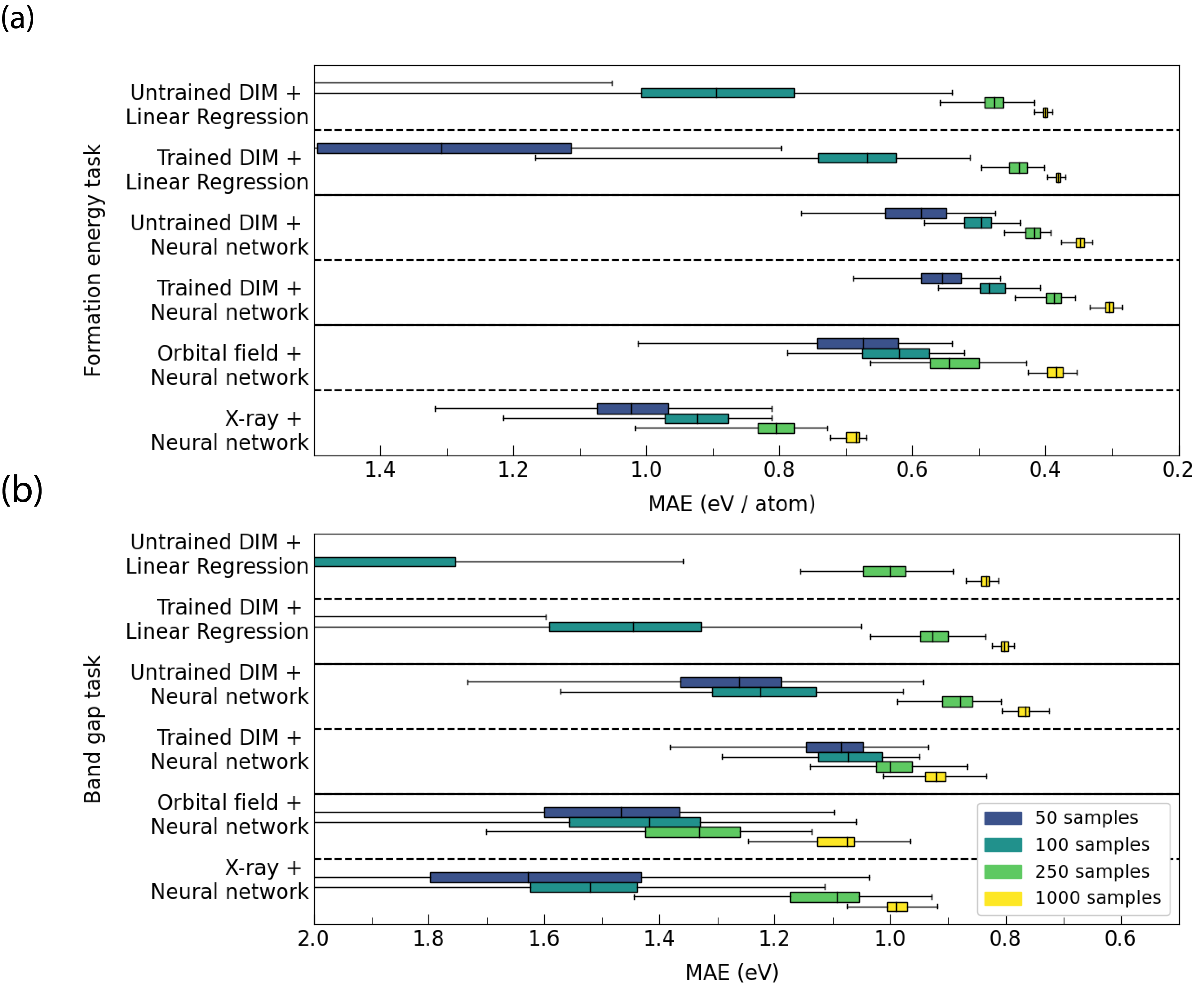}
	\caption[Box and whisker plots for the MAE on the test dataset are shown for supervised property prediction scikit-learn models trained on the band gap and formation energy tasks.]{
        Box and whisker plots for the MAE on the test dataset are shown for supervised property prediction scikit-learn models trained on the (a) formation energy and (b) band gap tasks. The results are shown as a function of the amount of training data (50, 100, 250 and 1000 samples), the representation provided to the model as input (untrained DIM, trained DIM, orbital field and x-ray), and the type of the regression model used (either a linear regression model or a neural network). Each box and whisker plot shows the data for 100 models, each trained on random subsamples of the training dataset, of which the Deep InfoMax model had full access to but without the property labels. The models trained were either linear regression models or scikit-learn neural networks with a single 64-node hidden layer using the ReLU activation function. The representations used were the normalised quasirandom feature vectors produced by an untrained Deep InfoMax model, a Deep InfoMax model trained on the full training dataset (either formation energy or band gap), and two engineered structural representations. The vectors produced by both the untrained and trained Deep InfoMax were of dimensionality 128. The engineered structural features used were the flattened orbital field matrix and the x-ray diffraction pattern from Matminer. The value added from Deep InfoMax training is dependent on both the task and the available data. Deep InfoMax training always results in an improvement in performance from the random baseline when training on the formation energy. For band gap, there is a significant improvement in performance associated with Deep InfoMax training for linear models and for neural networks with <= 100 available labels.
        }
	\label{fig:representation}
\end{figure*}

\subsubsection{Representation Learning}

The fixed global representation vectors produced by a Deep InfoMax model (length 128) have potential as input to traditional machine learning models. The trained Deep InfoMax model is used like any other featurisation function that creates a feature vector from the crystal structure for machine learning tasks. As a baseline to compare the learned representations with, we use the representation produced by an untrained / randomly weighted Deep InfoMax model and two feature vectors produced from Matminer featurisers. The vectors produced by the untrained model are an important point of comparison, as they demonstrate the value added of training over doing nothing. The vectors produced by the untrained model are normalised but the Deep InfoMax vectors are not. The two featurisers used allow Deep InfoMax to be compared with manual feature engineering. We chose the flattened orbital field matrix representation\cite{doi:10.1080/14686996.2017.1378060} and the x-ray diffraction pattern\cite{Cullity2001} because they are self-contained paramaterisations of the crystal.

We use scikit-learn to train the supervised downstream models at varying levels of property label availability (\cref{fig:representation}). The models chosen are least-squares-fit linear regression and a neural network with a single hidden layer with 64 nodes. The relative performance of the linear regression and a single hidden layer neural network are able to offer insights into the accessibility of the information in the representations. The linear models trained on the orbital field matrix and the X-ray diffraction pattern are not shown because the test MAEs for these models were several orders of magnitude greater than the other models. 

The Deep InfoMax representation results in a marked improvement compared to the untrained representation and Matminer features for the formation energy task. There is a marked improvement in performance using the Deep InfoMax representation over the untrained baselines across all data availability's for both the linear models and neural networks. For the band gap with small amounts of data ($<= 100$), Deep InfoMax improves linear models and makes downstream supervised model performance more consistent. The median performance is higher for neural networks trained using Deep InfoMax representations for small amounts of data, but the best performing models trained on the Deep InfoMax representations match the best performing models trained on the representations produced by a randomly weighted model. With greater amounts of data ($>= 250$), linear models continue to improve but the neural network trained on the Deep InfoMax representation fall behind both the untrained baseline and linear models trained on the Deep InfoMax representation. This suggests that with larger amounts of data ($>= 250$) the Deep InfoMax representation is more vulnerable to overtraining on the band gap task. The Deep InfoMax training regime aligns well with the formation energy task, but seems to require tuning to be more successful on band gap.


\begin{figure*}[!ht]
	\centering
	\includegraphics[width=\textwidth]{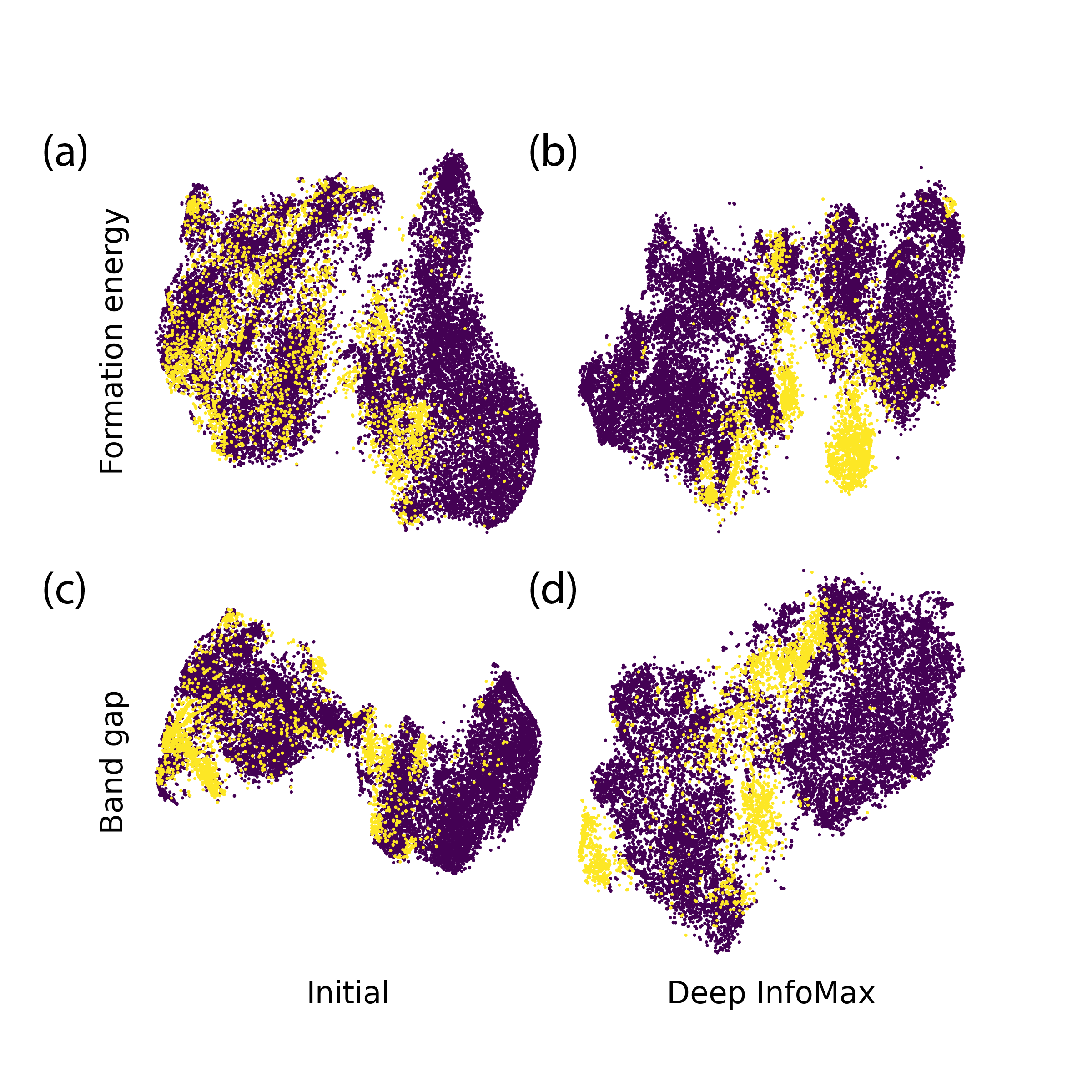}
	\caption[]{
        t-SNE plots demonstrating the distribution of halogen \& metal containing materials in the representation space are created using (b,d) trained Deep InfoMax representations and (a,c) representations produced from the ’initial’ model prior to Deep InfoMax training i.e random parameters. The t-SNEs are generated for the test partitions of the (a,b) formation energy and (c,d) band gap datasets. The t-SNEs are given an overlay depending on whether they contain both a metal and a halogen as a heuristic for identifying halogen salts. Materials that contain both a metal and a halogen are coloured yellow, while materials that do not meet these criteria are coloured purple. There is an increased clustering of materials containing both a halogen and a metal for both the band gap and formation energy latent spaces after training with Deep InfoMax. This kind of clustering based on simple chemical properties is an expected outcome of the training, as this kind of information is of direct relevance for identifying true and false samples.
	}
	\label{fig:haltsne}
\end{figure*}

\begin{figure*}[!ht]
	\centering
	\includegraphics[width=\textwidth]{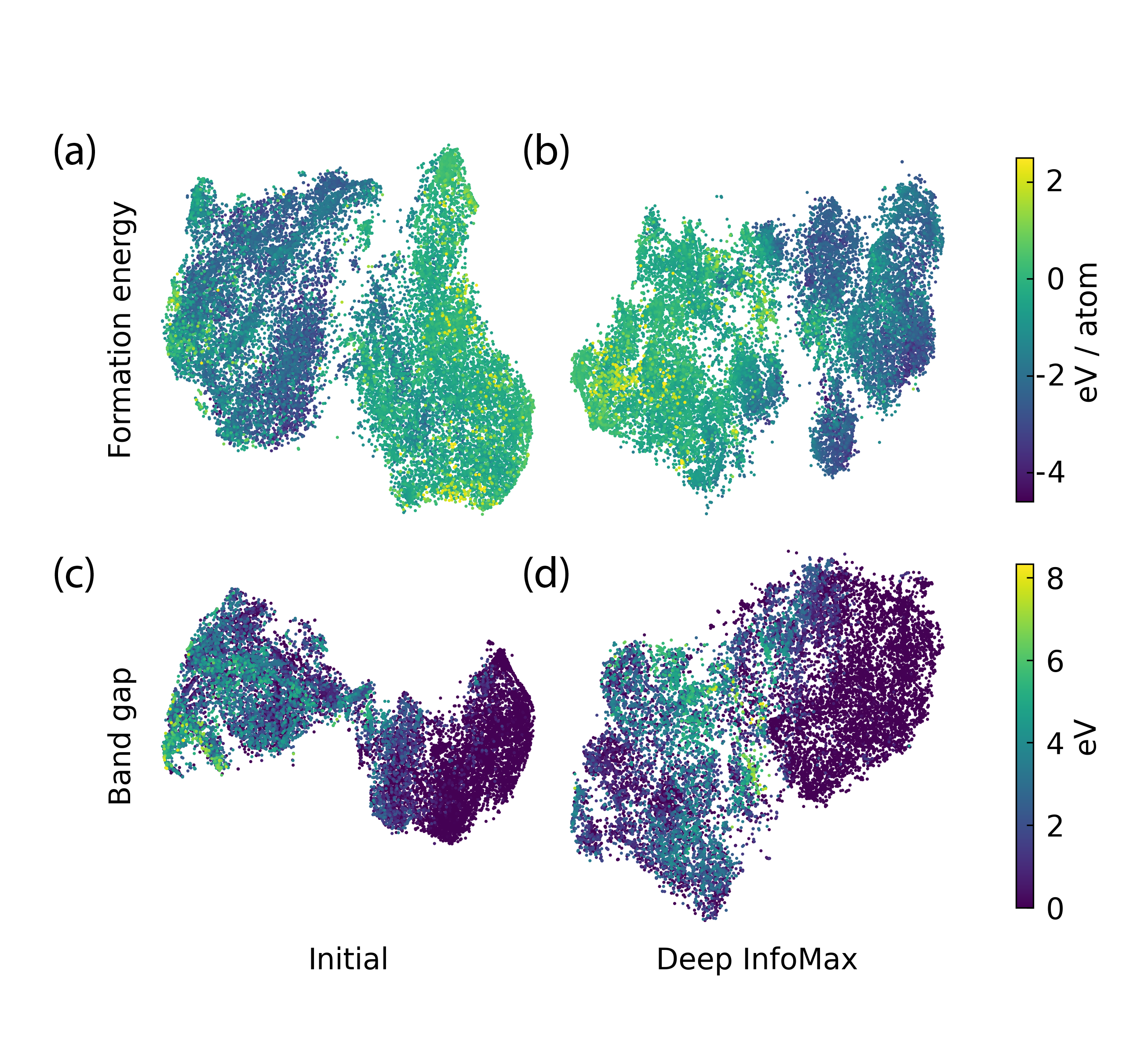}
	\caption[]{
      t-SNE plots highlighting the distribution of property labels in the representation space are created using (b,d) trained Deep InfoMax representations and (a,c) representations produced from the ’initial’ model prior to Deep InfoMax training i.e random parameters. The t-SNEs are generated for the test partitions of the (a,b) formation energy and (c,d) band gap datasets. The t-SNEs are given an overlay according to their relevant property label. Deep InfoMax training does not result in any obvious human readable differences in the distribution of the property labels. As Deep InfoMax is a self-supervised learning method with no direct access to property labels, there is no expectation that this would be the case.
	}
	\label{fig:proptsne}
\end{figure*}

\subsection{Visualisation Results}


To explore the properties of the global crystal representation learned by using Deep InfoMax, we generate t-SNE\cite{vanDerMaaten2008} plots to explore the topology of the representation. t-SNE is a dimensionality reduction technique that creates a two-dimensional representation of a high-dimensional space that preserves the nearest-neighbour distances for each point in the high-dimensional space. The global structure is not preserved in a t-SNE plot, so direct distances between points are only meaningful for nearby points in the t-SNE space. With these t-SNE plots it is possible to see how the materials cluster in the global feature space generated by self-supervised learning. The t-SNEs are generated on the test partitions of the band gap and formation energy datasets with scikit-learn using a perplexity of 100, or 5\% of the total number of data points. The other hyper parameters are set to scikit-learn defaults.

To investigate the topography of the latent space, t-SNEs are generated on top of the representations built by both an untrained, randomly initialised Site-Net model and a trained Deep InfoMax model for the test partitions of both the band gap and Formation Energy data sets. The perplexity is set to 100, or 5\% of the total number of data points. The other hyper parameters are set to scikit-learn defaults. t-SNEs with property labels as overlays (\cref{fig:proptsne}) contrast with t-SNEs overlayed with an innate property, the presence of a halogen and a metal (\cref{fig:haltsne}).  As the Deep InfoMax architecture is self-supervised, there is not much in the way of obvious clustering in the latent space in terms of property labels (\cref{fig:proptsne}). However, there is an increase in clustering when it comes to materials containing halogens and metals, especially with the formation energy data set, where halogen containing materials form an explicit "island" in the latent space (\cref{fig:haltsne}).

\begin{figure*}[!ht]
	\centering
	\includegraphics[width=1\textwidth]{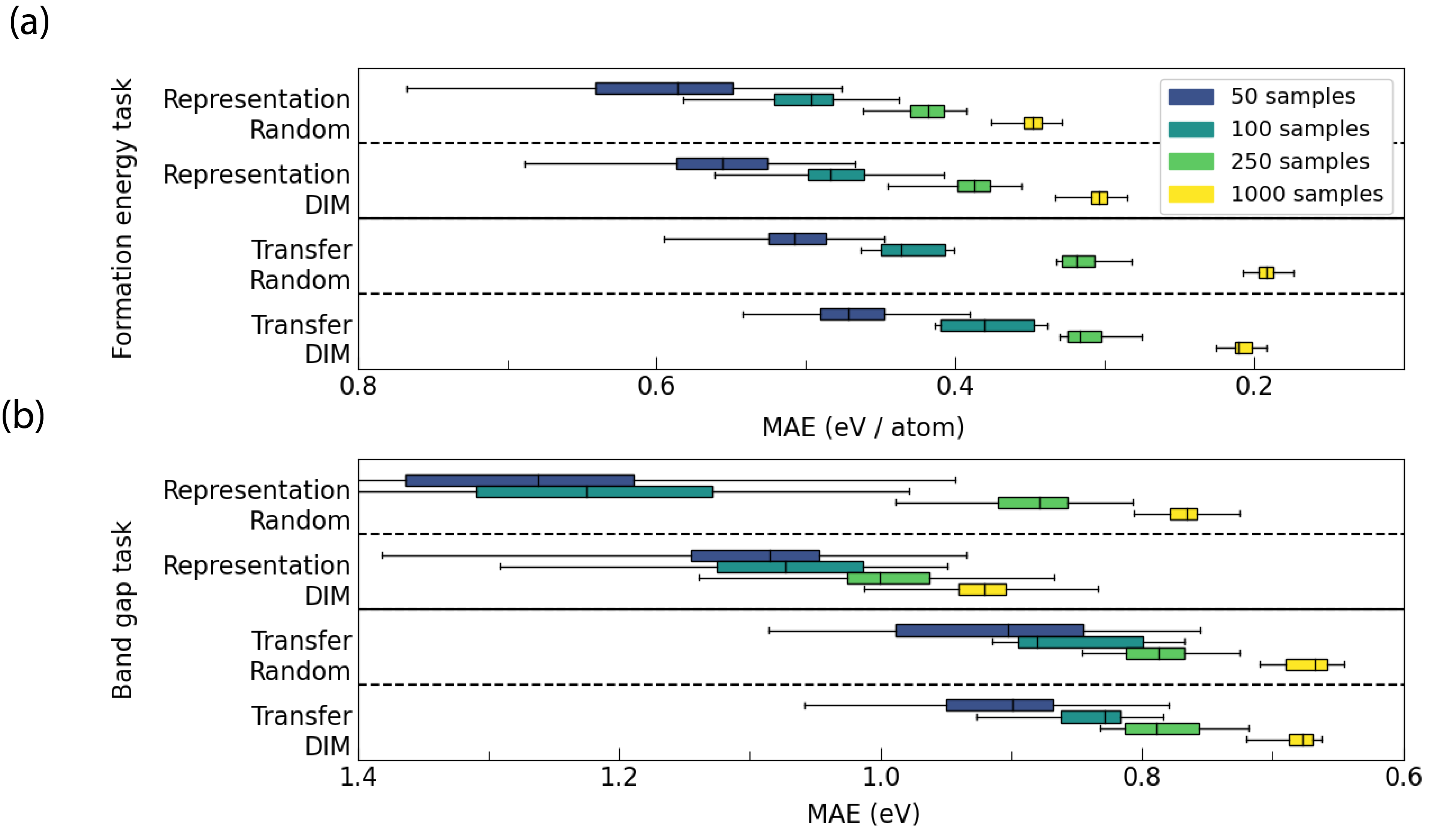}
	\caption[Training downstream neural networks on the representation produced by a trained Deep InfoMax (DIM) model and an untrained Deep InfoMax model with initial parameters (Representation) is compared to training supervised Site-Net models using both random starting parameters and starting parameters taken from a trained Deep InfoMax model (Site-Net).]
 {Training downstream neural networks on the representation produced by a trained Deep InfoMax (DIM) model and an untrained Deep InfoMax model with initial parameters (Representation, \cref{fig:representation}) is compared to training supervised Site-Net models using both random starting parameters and starting parameters taken from a trained Deep InfoMax model (Transfer, \cref{fig:transfer}). The methodologies are compared for the (a) formation energy task and (b) band gap task with label availability's varying from 50 to 1000. For representation learning, the box plots show 100 models, each trained on distinct randomly sampled property labels from the training dataset. For the Site-Net models, the box plots show 12 models using the same methodology. Results shown are the MAE of the models on the test dataset of $\sim$20,000 samples regardless of the amount of training data. Even with very small amounts of data, training a full Site-Net model results in superior performance for all cases. This is a surprising finding; since the Site-Net model has more than $10^5$ parameters and there are only 50 data points in the least data abundant case, we would expect severe overtraining and for training a small model on a frozen representation to be superior. This is not the case. The performance of Site-Nets with such small amounts of data suggests that this is an example of the double descent phenomena, where heavily over parameterised models eventually overcome overtraining and start to perform better than models optimised according to the bias-variance tradeoff. It is noted that both the Site-Net models and the downstream models trained in scikit-learn are using a single 64 node hidden layer in between the global representation and the prediction, and as such, they are directly comparable.
 }
	\label{fig:trans&rep}
\end{figure*}

\section{Discussion}

\subsection{Performance of Deep InfoMax for downstream learning}

There are consistent improvements in performance for pre-training with Deep InfoMax in both the representation and transfer learning contexts. In general, Deep InfoMax works best for the formation energy task rather than the band gap task, and works best when there are a small number of labels available for the supervised task. 

For representation learning (\cref{fig:representation}), Deep InfoMax is always the strongest representation in the small data regime (50-100 samples), with the median test MAE always lower than the representation from the untrained control model and the two Matminer featurisers. The Deep InfoMax features also result in more consistent downstream models for small data, with the variance in MAE being lower. This trend continues for higher amounts of training data (250-1000 samples) in the case of formation energy but falls apart for the band gap.

For transfer learning (\cref{fig:transfer}), results are more mixed. For formation energy, between 50 and 250 samples, starting training from the Deep InfoMax parameters gains consistent improvements in performance over starting with random parameters. This trend is reversed at 1000 samples where starting from random parameters results in a lower MAE. Transfer learning from Band Gap actively harms performance for all amounts of training data. For band gap, starting training from Deep InfoMax parameters results in little change to the MAE in either direction, with transfer learning from the formation energy being the strongest performer for all levels of data availability.

It is noted that the hyper parameters of the Deep InfoMax model were not explicitly optimised and were instead adjusted ad hoc from the original Site-Net hyper parameters until reasonable performance was obtained. There is further room for improvement in downstream performance through optimization of the hyper parameters of the Deep InfoMax model. With respect to Deep InfoMax's relative performance when compared to manual featurisers (\cref{fig:representation}) the performance of the hyper parameters is of importance. For the majority of the results, which compare Deep InfoMax training to untrained versions of the same model architecture, these comparisons are much less dependant on hyper parameters since they are shared between the models.

A surprising finding is that Site-Net models attain a better test MAE than small models trained on representations even with very small amounts of data (\cref{fig:trans&rep}). Given the hundreds of thousands of parameters on Site-Net, severe overtraining with small amounts of data would be expected. Despite this, the supervised Site-Net models using Deep InfoMax parameters as starting parameters (transfer learning) perform better than the downstream neural networks trained on top of the static representation produced by Deep InfoMax (representation learning) even with only 50 data points. This appears to be an example of the double descent phenomenon\cite{nakkiran2019deep} where heavily overparameterized models leave the region of overtraining and start to generalise again.

\begin{figure*}[!ht]
	\centering
	\includegraphics[width=1\textwidth]{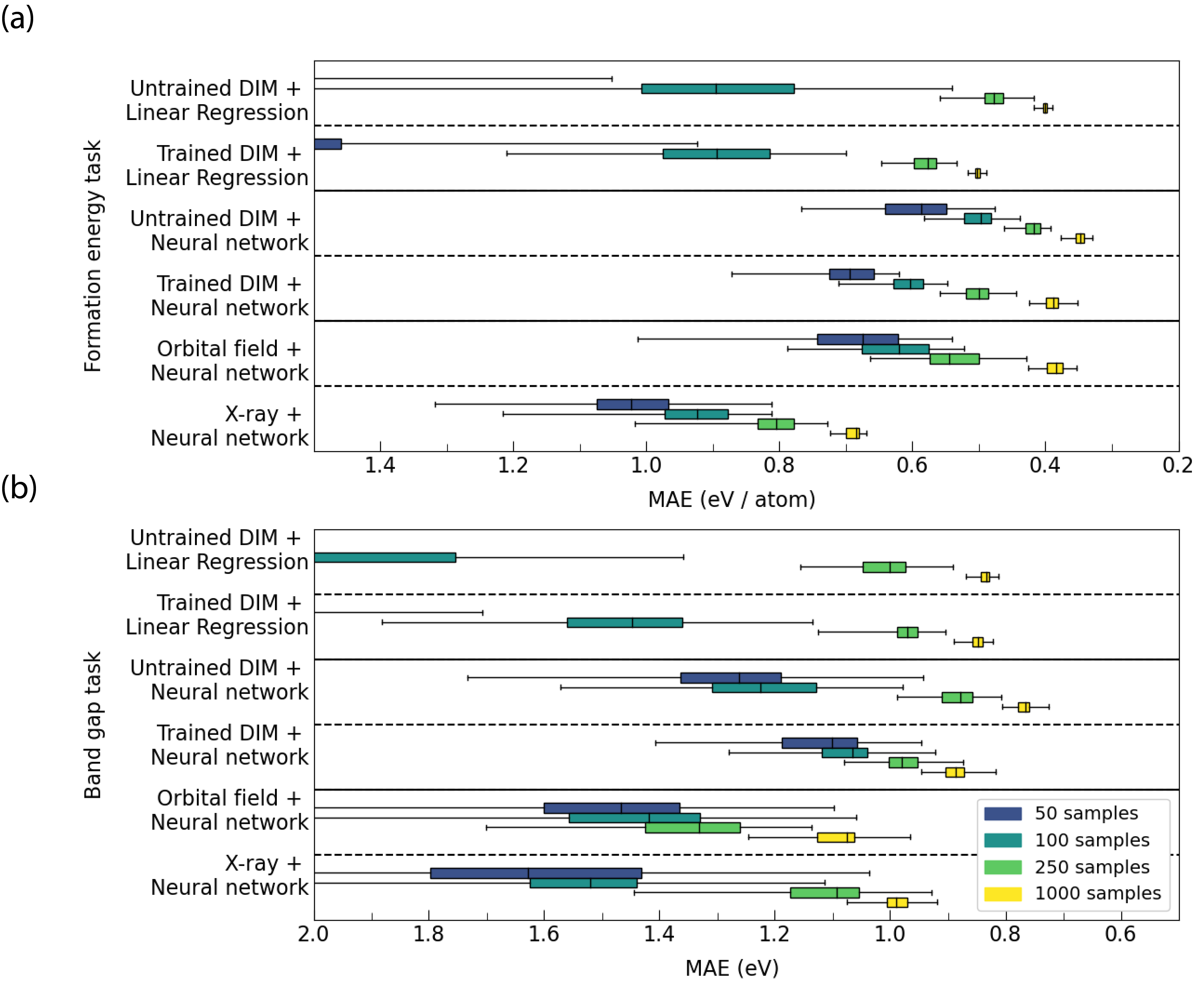}
	\caption[]{
      The representation learning experiment (\cref{fig:representation}) is performed using Deep InfoMax models without the additional false samples described in (\cref{fig:False_Samples}) for the (a) formation energy and (b) band gap tasks. Without the synthetic false samples, training using Deep InfoMax representations results in uniformly worse performance for the formation energy task. For the band gap the performance of Deep InfoMax is similar except that the improvement in the performance of linear models is no longer present. The false sampling strategy chosen in this work had a greater impact on the formation energy task than the band gap task, suggesting that fine tuning the false sampling strategy is specific to the downstream task. An alternative set of synthetic false samples tuned to the band structure would likely result in a similar improvement in the performance for band gap.
	}
	\label{fig:controlrep}
\end{figure*}

\subsection{Domain knowledge applications in false sampling}

The superior performance of Deep InfoMax for the formation energy task when compared to the band gap task is likely due to the chosen enhancements to the false samples. Without augmenting false samples, the formation energy performs poorly when used for representation learning, worse than the untrained control in all cases (\ref{fig:controlrep}). The addition of the engineered false samples makes the difference in the learned representations being superior to random baselines for formation energy. The current engineered false samples are orientated towards the correctness of the local environment in real space, which aligns with the formation energy task because formation energy is primarily based on the geometry of the crystal and the local charge environment at each site. The band structure, and therefore the band gap, is more complicated. Performance on the band gap task is likely to be improved through an alternative false sampling scheme more focused on preserving information about the band structure.

Prioritising speed, the false samples used in this implementation of Deep InfoMax relied on data donation from other crystals in the dataset, which is chosen because it both provides a large sample space (proportional to the square of the size of the dataset) and the operations can be trivially performed on the GPU during training. These false samples are also conceptually simple and not tuned for any particular property though they empirically align better with the formation energy task than the band gap task. An alternative to the data manipulation approach to false sampling would be to produce false samples prior to, or during training through chemical transformations applied to the crystal structure. This would allow the model to consider defect chemistry, or more chemically complex perturbations of the lattice. For example, to improve the performance of downstream learning on the band gap, a false sampling strategy that focuses on dopants and other small perturbations to the crystal that are known to make large changes to a material band structure might be considered. The false sampling methodology is an important hyper parameter of the Deep InfoMax model that is heavily dependant on domain knowledge with respect to the target supervised task.

\subsection{Untrained models are an essential, non-trivial baseline}

A noteworthy result in the experiments with representation learning is the representation produced by untrained Deep InfoMax models being stronger than the engineered structural features when there are small amounts of data, in addition to the untrained representation being generally competitive with the trained Deep InfoMax models. There is precedent to this result in other areas.

In recent work in NLP, it has been found explicitly that untrained transformers sometimes make stronger sentence encoding than trained transformers\cite{wieting2019training}, suggesting that the pretrained word embeddings are the key determinant of performance. Random transformers, using what is essentially random weighted sums of random projections are sufficient to create a representation if the word embeddings are good. It is not surprising that this result is also found with crystal transformers, as elemental embeddings are very similar to word embeddings with a vocabulary defined by the periodic table and a grammar defined by the structure of the crystal. In the domain of materials science, random projections have been shown to often outperform compositional feature engineering for composition based property prediction tasks\cite{D2DD00039C}.

From the result obtained in this work and precedent in other domains, the untrained versions of a self-supervised models for crystals are demonstrated to be a natural baseline for establishing the efficacy of the training methodology. It is therefore a recommendation that future work in self-supervised learning for materials informatics adopt untrained versions of models as baselines. Demonstrating that trained models achieve superior performance to untrained models is essential, and shown not to be something that can be assumed even if performance is good.

\subsection{Practical application of Deep InfoMax in materials informatics and compatibility with other models}

This work represents a controlled pilot investigation into the application of Deep InfoMax within materials informatics. By training Deep InfoMax on large supervised datasets and performing downstream training on sub samples of the same dataset we ensure that the distribution of inputs learned by Deep InfoMax is the same as the distribution in the supervised sub samples. In addition, by resampling the supervised subsets and repeating experiments, we are able to obtain a well-defined statistical resolution on the performance differences between approaches. Removing distributional shift allows the separation of the value of Deep InfoMax from impact of the distributional shift between datasets. Another way to phrase this, is that the problem that Deep InfoMax is being applied to in this work is pure interpolation with no extrapolation.

In practical applications of Deep InfoMax distributional shift is actively desirable, and learning about classes of material not present in the small supervised dataset is a key source of value. If there are materials in the pretraining data that are "out of distribution" with respect to the supervised data. This has the potential to increase the probability that the downstream supervised model will be able to extrapolate to these cases as they are already accounted for during pre-training. Pre-training on a large dataset of materials such as the ICSD\cite{Hellenbrandt2004} can give the model knowledge of materials that are very different from those present in a small supervised dataset. Through the demonstration of benefits from Deep InfoMax pretraining in the in-distribution case. This sets the necessary precedent for use of Deep InfoMax for novel extrapolatory cases.

The Deep InfoMax loss function described in this work can be integrated into other architectures quite easily. An autoencoder reconstruction loss, a contrastive loss, and the Deep InfoMax loss can be trivially combined in the same model. This ease of combination comes from them all sharing the encoder part of the architecture, and simply being different branching "decoders" than can be forked from a common bottleneck. Combining the approaches together and weighting the loss functions may work better than any one method on its own, with each loss function focusing on different elements of a good representation. Equivalently, Deep InfoMax can be integrated into a supervised learning pipeline and the loss function can co-exist with a supervised loss.  Contrastive learning in particular is a promising candidate for combination with Deep InfoMax, as a technique similar to Deep InfoMax which has seen recent success in materials informatics\cite{ottomano2023enhancing,D2CC01764D,Magar2022a}. Both contrastive learning and Deep InfoMax employ the idea of "positive pairs" and "negative pairs", and manipulate the representation of the material so that positive pairs belong together and negative pairs do not. The key difference in the approaches is that contrastive learning operates by defining positive pairs and negative pairs of complete crystals and complete local environments. By contrast, Deep InfoMax  compares complete crystals and complete local environments to the building blocks they are made up of.

\section{Conclusion}

We introduce the use of Deep InfoMax for self-supervised learning in material informatics through application to the Site-Net architecture for the purposes of transfer learning and representation learning. Deep InfoMax first encodes the crystals to fixed sized vectors in a shared latent space, then uses a classifier to verify the quality of the representation by differentiating 'true' local samples that belong to the relevant crystal from 'false' samples which do not. In our implementation, we introduce an additional degree of freedom to the Deep InfoMax methodology, where domain knowledge is used to curate and bias the false samples such that the information preserved in the representation is more relevant to downstream tasks.

We benchmark the suitability of Deep InfoMax through self-supervised pretraining, where performance on supervised datasets with small numbers of available labels is improved through self-supervised training on large datasets without property labels. We then transfer that knowledge to supervised training on smaller datasets where property labels are available. To benchmark Deep InfoMax independently of distributional shift, we pretrain on the full supervised dataset treating it as unlabelled and then train supervised models on randomised subsets of the data. We robustly show that Deep InfoMax pretraining has value added for improving performance in this interpolative regime. We also demonstrate that performance improves when an appropriate false sampling strategy used for a given supervised task, showing a direction for future improvement based on domain knowledge.

In this work, Deep InfoMax was applied to the Site-Net architecture, however, the Deep InfoMax loss function can be used with other architectures such as CGCNN, ALIGNN, and MegNet\cite{cgcnn,Choudhary2021,megnet} that operate on crystal structures. In addition, Deep InfoMax is trivial to combine with other loss functions as part of larger self-supervised learning frameworks. We present Deep InfoMax, adapted to the domain of materials informatics, as a promising candidate for inclusion in foundation models as well as having general applicability as a self-supervised learning technique for materials science.

\section{Author Contributions}

Michael Moran: Conceptualisation, Data Curation, Formal Analysis, Investigation, Methodology, Software, Visualization, Validation, Writing – original draft, Writing – review \& editing
Michael Gaultois: Conceptualisation, Formal Analysis, Methodology, Visualization, Supervision, Writing – review \& editing
Vladamir Gusev: Conceptualisation, Formal Analysis, Methodology, Supervision, Writing – review \& editing
Matthew Rosseinsky: Supervision, Writing – review \& editing
Dmytro Antypov: Supervision, Writing – review \& editing

\section{Code availability}
The code developed for this work is available at https://github.com/lrcfmd/Site-Net-DeepInfoMax

\section{Conflicts of interest}
There are no competing interests to declare.

\section{Acknowledgements}
\label{sec:others}
Work was performed using Barkla, part of the High Performance Computing facilities at the University of Liverpool, UK. The authors thank the Leverhulme Trust for funding via the Leverhulme Research Centre for Functional Materials Design. MWG thanks the Ramsay Memorial Fellowships Trust for funding through a Ramsay Trust Memorial Fellowship.




\bibliographystyle{unsrtnat}
\bibliography{references} 





\end{document}